\documentclass[letterpaper, 10 pt, conference]{ieeeconf}  

\IEEEoverridecommandlockouts                              

\usepackage{xcolor}

\newcommand{\model}{Trinity-Net}

\newcommand{\etal}{\textit{et al.}}
\newcommand{\eg}{e.g.,~}

\usepackage[acronym]{glossaries}

\let\labelindent\relax
\usepackage[shortlabels,inline]{enumitem}

\usepackage{multirow}
\usepackage{booktabs}

\setlist[enumerate]{noitemsep,leftmargin=*,topsep=0em}

\usepackage{hyperref}

\hypersetup{%
  colorlinks=true,%
  linkcolor={red!50!black},
  citecolor={blue!65!black},
  urlcolor={blue!80!black},
  bookmarksnumbered=true,%
  bookmarksopen=true}

\usepackage[capitalise]{cleveref}

\newacronym{SAM}{SAM}{Segment Anything Model}
\newacronym{CNN}{CNN}{Convolutional Neural Network}
\newacronym{OOD}{OOD}{Out-Of-Distribution}
\newacronym{mIoU}{mIoU}{mean Intersection over Union}
\newacronym{IoU}{IoU}{Intersection over Union}
\newacronym{resR}{resR}{Residual Recall}

\newdimen\TreeIndent
\def\Branch#1{%
  \par\noindent
  \hskip\TreeIndent
  \vrule width0.4pt height1.6ex depth0pt
  \raise0.7ex\hbox{%
    \vrule width0.5em height0.4pt depth0pt
  }%
  \hskip0.4em
  #1%
}

\def\LastBranch#1{%
  \par\noindent
  \hskip\TreeIndent
  \raise0.7ex\hbox{%
    \vrule width0.4pt height.9ex depth0pt
  }%
  \raise0.7ex\hbox{%
    \vrule width0.5em height0.4pt depth0pt
  }%
  \hskip0.4em
  #1%
}

\usepackage{siunitx}
\usepackage{arydshln}

\makeglossaries

\title{\LARGE \bf
Trinity: Unifying Class-Agnostic Terrain and Semantic Segmentation for Unstructured Outdoor Environments by Leveraging Synthetic Data
}

\author{Marcus G Müller$^{1}$$^{2}$$^{*}$, Wout Boerdijk$^{1}$, Maximilian Durner$^{1}$, Riccardo Giubilato$^{1}$, Abel Gawel$^{3}$, Wolfgang Stürzl$^{1}$, \\Roland Siegwart$^{2}$, Rudolph Triebel$^{1}$
\thanks{${*}$ \ Corresponding author {\tt\small marcus.mueller@dlr.de}}
\thanks{${**}$This work was supported by the Helmholtz Association project iFOODis (contract number KA2-HSC-06)}
\thanks{$^{1}$Institute of Robotics and Mechatronics, German Aerospace Center (DLR), Weßling, Germany}
\thanks{$^{2}$Federal Institute of Technology Zurich (ETH Zurich), Zurich, Switzerland}
\thanks{$^{3}$Robotics and AI Institute (RAI), Zurich, Switzerland}
}

\usepackage{bbm}
\usepackage{relsize}
\usepackage{multirow}
\usepackage{graphicx}
\usepackage{float}
\usepackage{adjustbox}
\usepackage{amssymb}
\usepackage{stfloats} 

\begin{document}

\maketitle
\thispagestyle{empty}
\pagestyle{empty}

\begin{abstract}
Terrain understanding is fundamental for mobile robots operating in unstructured outdoor environments. Existing vision-based traversability estimation methods rely on robot-specific annotations or semantic class mappings, limiting transferability across platforms and requiring costly re-annotation when robot capabilities change, while standard semantic segmentation methods only focus on specific predefined classes, which do not capture the variety of terrains. In this work, we propose a transformer-based architecture that jointly performs class-specific semantic segmentation and class-agnostic terrain segmentation within a unified network, called Trinity.
Terrain regions are segmented based solely on visual appearance, without predefined semantic labels or robot-dependent traversability scores. This formulation enables the learning of robot-agnostic visual terrain priors that can be combined with robot-specific experience for downstream tasks such as traversability estimation, visual odometry, and mission planning. 
To enable large-scale training with diverse terrain appearances, we extend the OAISYS simulator and introduce RUGDSynth, a synthetic dataset inspired by RUGD with class-agnostic terrain samples. Furthermore, we present the EXTerra Dataset, providing real-world images annotated with both class-specific and class-agnostic terrain labels. Experiments demonstrate the feasibility of the proposed task and the effectiveness of our joint segmentation approach in complex outdoor environments.
Code and datasets will be released with this publication (after review).
\end{abstract}

\section{Introduction}
Semantic scene understanding is a fundamental capability for mobile robots operating in complex outdoor environments.
In such settings, reliable perception supports navigation, state estimation, path planning, and manipulation.
It can further enable higher-level objectives such as identifying scientifically relevant areas or suitable terrain for infrastructure deployment in planetary exploration.

A central component of outdoor scene understanding is terrain segmentation, with traversability estimation as a closely related task. 
While supervised semantic segmentation has achieved remarkable success in structured urban environments~\cite{van_brummelen_autonomous_2018}, transferring these approaches to natural terrain remains challenging. 
In contrast to roads or buildings with well-defined geometry, outdoor environments exhibit irregular regions, gradual transitions, and visually similar surface types. 
Boundaries are often ambiguous, and semantic categories are less clearly separable, making also the annotation process difficult~\cite{swan_ai4mars_2021}.

\begin{figure}[!tp]
\centering
\vspace{0.3em}
\includegraphics[width=\linewidth]{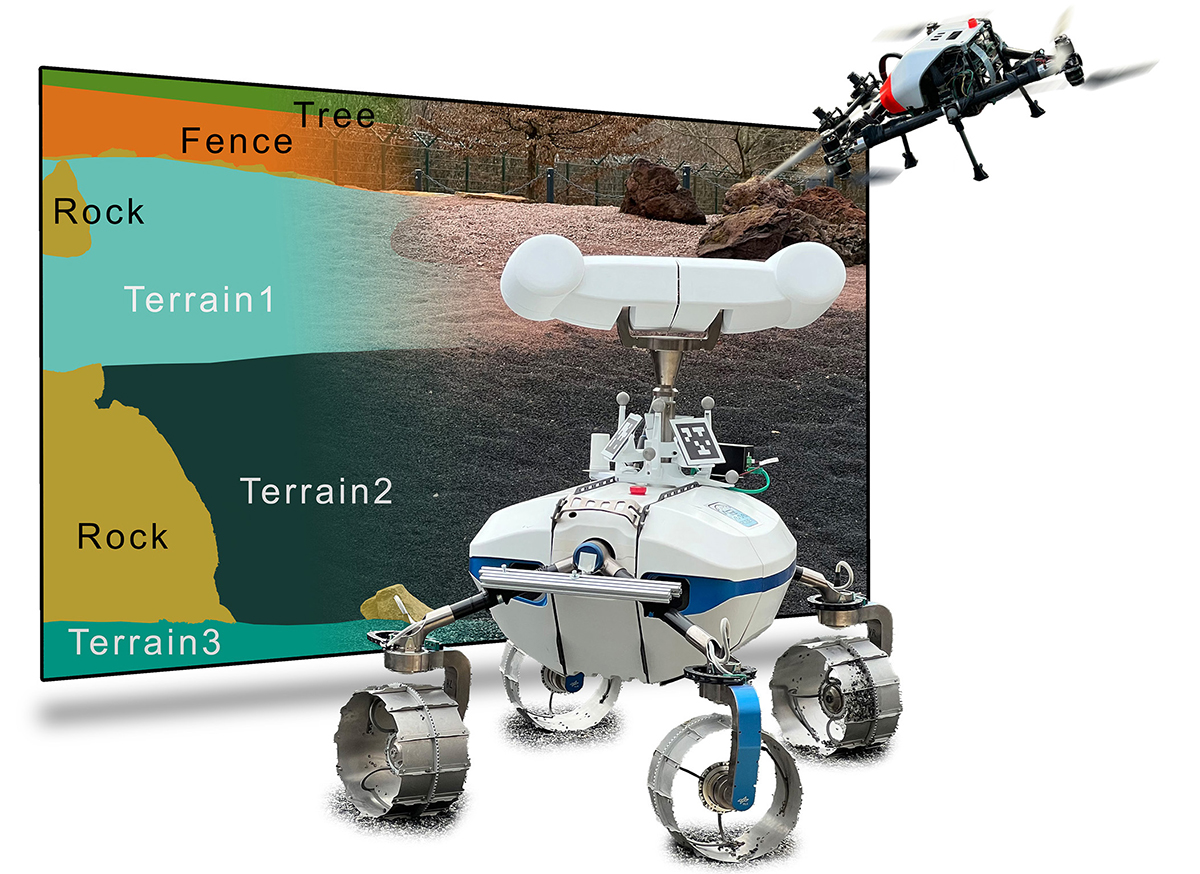}
\vspace{-1.7em}
\caption{A rover and a drone observe a planetary exploration outdoor laboratory. The overlapping annotations shown on the color image illustrate the ideal segmentation results produced by the proposed method. The task is to segment different terrain types (shown in shades of cyan) in a class-agnostic manner, enabling their use across a wide range of robotic applications and platforms, while simultaneously identifying class-specific regions that are important for a mission (here: rock, fence, and tree).}
\label{fig:lru_outdoor}
\vspace{-1.2em}
\end{figure}

Beyond these perceptual challenges, terrain semantics are inherently context-dependent.
The appearance and meaning of concepts such as soil or grass vary with environmental conditions (\eg weather, season, illumination) and with the operational context of the robot. 
A compact dirt path may be easily traversable for a wheeled platform in dry conditions, yet become impassable mud after rainfall; dense grass may pose no difficulty for a tracked vehicle but obstruct a small wheeled robot. 
Consequently, semantic labels and especially traversability assessments are not absolute properties, but depend on both environmental and robot-specific factors.
This variability makes it difficult to define a universally valid, fixed taxonomy of terrain classes.

Most existing methods, however, assume precisely such predefined class sets and operate under a closed-world assumption in which all relevant categories are known during training. 
In traversability estimation, supervision is further tied to a specific robot platform, since annotations implicitly encode its mobility capabilities. 
As a result, deploying models in new environments or on different robots typically requires new data collection and retraining, limiting scalability and robustness.

To address these challenges, we introduce \model, a unified transformer-based architecture for flexible scene understanding that jointly performs class-agnostic and class-specific pixel-wise segmentation without relying on fixed terrain taxonomy.

The class-agnostic part separates scenes into visually coherent terrain regions without predefined semantic labels. 
By focusing purely on appearance, 
it provides a robot-agnostic terrain representation that can serve as a visual prior for (self-supervised) downstream tasks~\cite{frey_fast_2023,jung_v-strong_2024}.
Unlike local region-proposal methods~\cite{kirillov_segment_2023}, \model~is able to enforce global visual consistency across spatially disjoint terrain regions.

The class-specific branch predicts platform- and context-independent or mission critical semantic categories (\eg trees, rocks, buildings). 
This separation allows the model to retain stable semantic anchors while avoiding an overconstrained terrain taxonomy. \cref{fig:lru_outdoor} gives an illustration of the overall task.

Training such a system requires substantial terrain diversity to ensure robust generalization and genuinely class-agnostic behavior.
However, annotating large-scale datasets with terrain labels is costly and time-consuming.
To address this limitation, we additionally leverage synthetic data.
Unfortunately, many existing outdoor datasets do not provide terrain-specific annotations.
Moreover, most simulation environments are unable to output such information, as their semantic labels are typically object-based rather than texture- or material-based.
In contrast, the OAISYS simulator~\cite{muller_photorealistic_2021} provides semantic annotations using a material-based labeling approach. This enables the extraction of terrain annotations corresponding to different surface textures.

To this end, we leverage and extend the OAISYS simulator to generate a large-scale synthetic dataset for joint training. Our approach achieves strong performance on the RUGD benchmark~\cite{wigness_rugd_2019} across both segmentation tasks. Furthermore, we demonstrate transferability by evaluating the trained model—developed in a field robotics context—on a real-world dataset recorded for planetary exploration scenarios.

To summarize, our contributions are:
\begin{enumerate*}[label=(\roman*)]
    \item \textbf{\model, a unified transformer architecture} for joint class-specific semantic and class-agnostic segmentation, trained on synthetic data to learn transferable appearance-based terrain representations;
    \item an \textbf{extension of the OAISYS simulator} enabling large-scale synthetic data generation for this task, resulting in
    \textbf{RUGDSynth}, a synthetic counterpart to the field-robotics RUGD benchmark, and
    \item the \textbf{EXTerra Dataset}, a real-world planetary exploration dataset from an analog mission site.
\end{enumerate*}

\section{Related Work}
Our work builds on semantic segmentation~\cite{csurka_semantic_2022,thisanke_semantic_2023}, with emphasis on terrain segmentation, traversability estimation, and open-world segmentation.

\textbf{Terrain Segmentation}
is commonly formulated as supervised semantic segmentation with predefined terrain taxonomies, requiring extensive manual annotation and fixed class definitions.
These limitations become particularly evident in planetary exploration, where annotated data is scarce and highly variable terrain appearance challenges rigid semantic categories.
Existing approaches range from rock-focused obstacle classification~\cite{durner_autonomous_2023,liu_rockformer_2023} to broader supervised semantic labeling efforts~\cite{swan_ai4mars_2021,gonzalez_deepterramechanics_2018}.
To improve robustness multimodal perception is additionally explored, by fusing visual and thermal imagery~\cite{castilla_arquillo_omniunet_2025}.
Several works also investigate reduced-supervision,
including sparse labeling~\cite{goh_mars_2022}, semi-supervised learning~\cite{zhang_smars_2024}, and text-guided foundation-model-based segmentation~\cite{fang_mtsnet_2024}. 
Despite mitigating annotation scarcity, these methods still operate within predefined terrain categories.
Similar reliance on supervision and fixed semantic taxonomies appears in terrestrial unstructured outdoor environments.
For instance, CRLNet~\cite{li_crlnet_2026} combines convolutional and attention mechanisms in a supervised setting, reporting strong results on RUGD~\cite{wigness_rugd_2019} and RELLIS-3D~\cite{jiang_rellis-3d_2021}.
Other approaches reduce semantic labels to navigability-oriented categories.
In~\cite{guan_ga-nav_2022} a multi-scale transformer to aggregate contextual features and cluster terrains into roughness-based groups for navigation is employed. 
Similarly, Li~\etal~\cite{li_contextual-aware_2025} propose a contextual-aware segmentation network that models local and global dependencies while operating within the same roughness-based terrain taxonomy.
Closely related to our motivation, Ellis~\etal~\cite{ellis_temporally_2025} introduce a temporally consistent unsupervised approach for class-agnostic terrain segmentation.
While they only address class-agnostic region discovery, we additionally support class-specific semantic segmentation within a single framework.

\textbf{Traversability Estimation.}
While terrain segmentation assigns semantic labels to surface regions, traversability estimation models navigability, commonly formulated as a binary decision or continuous feasibility score.
Schilling~\etal~\cite{schilling_geometric_2017} combine visual and geometric features using random forest.
Other methods use elevation maps to predict traversability via learning-based approaches~\cite{chavez-garcia_learning_2018,yang_real-time_2021}.
In~\cite{shaban_semantic_2022} dense traversability maps from sparse point clouds for off-road navigation are learned.
Several approaches exploit self-supervision from robot interaction.
Kahn~\etal~\cite{kahn_badgr_2021} cast traversability as an end-to-end reinforcement learning problem.
Other methods leverage platform-dependent (weak) supervision, such as foothold reprojection for legged robots~\cite{wellhausen_where_2019}, 
MPC trajectories~\cite{gasparino_wayfast_2022},
or proprioceptive signals including odometry and IMU signals~\cite{sathyamoorthy_terrapn_2022}. 
Building on these ideas combined with foundation-model representations, Frey~\etal~\cite{frey_fast_2023} propose an online self-supervised system that adapts their model using visual embeddings and short human demonstrations, while Jung~\etal~\cite{jung_v-strong_2024} incorporate segmentation priors using \gls{SAM}~\cite{kirillov_segment_2023} to improve fine-grained traversability prediction.

\textbf{Open-World Segmentation}
moves beyond fixed categories and closed-set assumptions to handle previously unseen categories during execution. 
Some works estimate predictive uncertainty to detect unknown regions, through Bayesian inference~\cite{sapkota_bayesian_2022,muller_uncertainty_2023}.
Further, auxiliary datasets are leveraged to explicitly expose models to unknown samples during training~\cite{chan_entropy_2021,bevandic_simultaneous_2019}.
Generative~\cite{lis_detecting_2024,zhao_omnial_2023} and reconstruction-based~\cite{bergmann_uninformed_2020,zhang_destseg_2023} methods detect novel regions by identifying discrepancies between input images and reconstructed outputs, as unseen patterns typically yield higher reconstruction errors. 
In \cite{tsai_multi-scale_2022} distance-based unsupervised anomaly detection is introduced.
Finally, feature-space approaches aim to separate known from unknown classes by measuring distances to learned class clusters within the feature space~\cite{sodano_open-world_2024,blum_scim_2023}.

\section{Class-Specific and Class-Agnostic Ground Segmentation}
We introduce \model, a novel unified framework for class-agnostic terrain and class-specific semantic segmentation, enabled by a large synthetic dataset spanning diverse terrains.
It is composed of three main parts, hence the name of the model.
We first present the architecture, followed by the simulation and dataset generation pipeline.

\subsection{Architecture - \model}

\begin{figure*}[th!]
\centering
\vspace{1.5em}
\includegraphics[width=\textwidth]{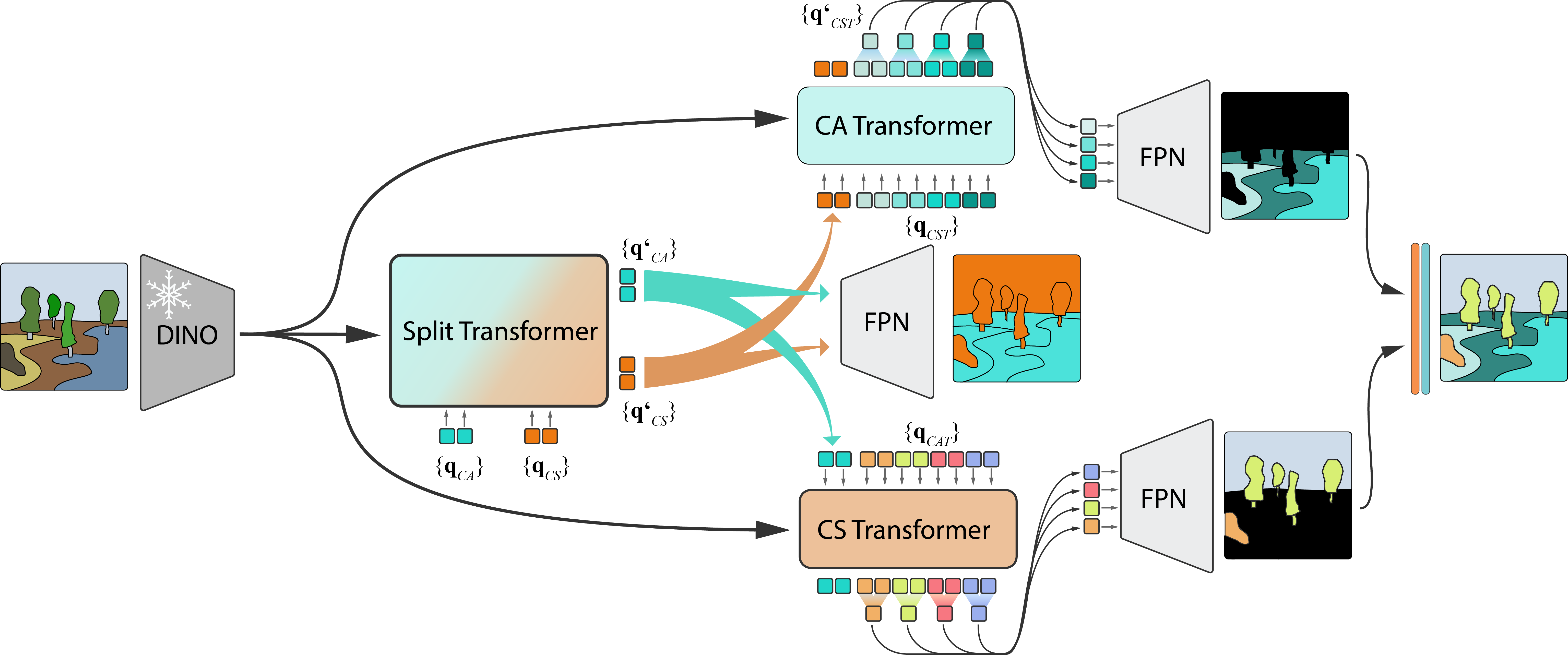}
\vspace{-1.5em}
\caption{The figure provides an overview of \model. The model consists of an image encoder and three main components: the Split Transformer, the CS Transformer, and the CA Transformer. The input image is first processed by a DINOv2 image encoder to obtain a feature map. Queries in the Split Transformer attend to the image features and implicitly divide the feature space into a CS and a CA component. The resulting queries that attend to the CS component are extended to form the queries of the CA Transformer, while the queries attending to the CA component are expanded to form the queries of the CS Transformer. The queries of both the CS and CA Transformers attend to the feature map. The resulting output queries are then aggregated and upsampled. Finally, the class-specific output is concatenated with the class-agnostic output.}
\label{fig:trinity_net}
\vspace{-1.2em}
\end{figure*}

Our method addresses two related but distinct sub-tasks: 
\textit{class-specific (CS) semantic segmentation} of predefined categories known a priori, capturing platform-independent and mission-relevant semantics, and \textit{class-agnostic (CA) terrain segmentation} that groups visually coherent ground regions in a robot- and context-agnostic manner.
Each pixel $p$ is assigned exclusively to either a specific class $C_{CS}$ or to a non-specified ground class $C_{CA}$:
\begin{equation}
p \in C_{CS} \cup C_{CA}
\;\;\; \textrm{where}\ \;
C_{CS} \cap C_{CA} = \varnothing.
\label{eq:cs_ca_eq}
\end{equation}

The overall architecture is illustrated in \cref{fig:trinity_net}.
We first encode the input RGB image
using a frozen DINOv2 backbone~\cite{oquab2024dinov2learningrobustvisual}.
The extracted image features
are passed to a DETR-like \textit{Split-Transformer} with learnable query sets, 
$\{\mathbf{q}_{CS}\}$ and $\{\mathbf{q}_{CA}\}$,
which partition the feature space into class-specific and class-agnostic regions.
To explicitly encourage the intended partitioning of the feature space, we apply an auxiliary loss to the updated query embeddings. 
For this purpose, we map the ground-truth annotations to a binary representation distinguishing CS from CA regions.

The updated querys $\{\mathbf{q'}_{CS}\}$ and $\{\mathbf{q'}_{CA}\}$ are then separated and passed to two subsequent transformer modules:
the \mbox{\textit{CS-Transformer (CST)}}, responsible for segmenting class-specific semantics, and the \mbox{\textit{CA-Transformer (CAT)}}, which handles class-agnostic regions.
Each transformer operates with its own set of learnable queries, $\{\mathbf{q}_{CST}\}$ and $\{\mathbf{q}_{CAT}\}$, representing prototypes for the respective regions.
Since a single query cannot capture the full complexity of a CA region or CS category in the real-world, we employ multiple queries per region to represent its variability and increase robustness.

To guide each transformer toward its relevant regions, we expand the task-specific queries with the updated queries:
$\{\mathbf{q'}_{CA}\} \cup \{\mathbf{q}_{CST}\}$ and $\{\mathbf{q'}_{CS}\} \cup \{\mathbf{q}_{CAT}\}$.
Intuitively, the Split-Transformer attends to complementary regions of the scene for each task, thereby providing contextual information that facilitates separation.

We average the resulting task-specific queries, $\{\mathbf{q'}_{CST}\}$ and $\{\mathbf{q'}_{CAT}\}$, that belong to the same region.
This aggregation step reduces the queries to match the final CS and CA output channels and is performed prior to upsampling for memory efficiency.
Note, the number of CA prototypes is chosen to be sufficiently large to accommodate diverse ground configurations.
The amount of final CS output channels depends on the mission-relevant semantic classes.
Finally, the aggregated queries are subsequently upsampled and concatenated to form a unified prediction tensor containing both class-specific and class-agnostic segmentation maps.

Since the CAT does not enforce a fixed alignment between queries and regions, the same terrain may be represented by different queries across samples. 
To resolve this ambiguity, predictions are matched to ground-truth masks using a Hungarian matcher to determine the optimal alignment between predicted class-agnostic regions and ground-truth masks.
After this matching step, a cross-entropy loss is applied to the full network output.

\subsection{Data Generation - RUGDSynth}

\begin{figure*}[!b]
\centering
\vspace{-1.2em}
\includegraphics[width=\textwidth]{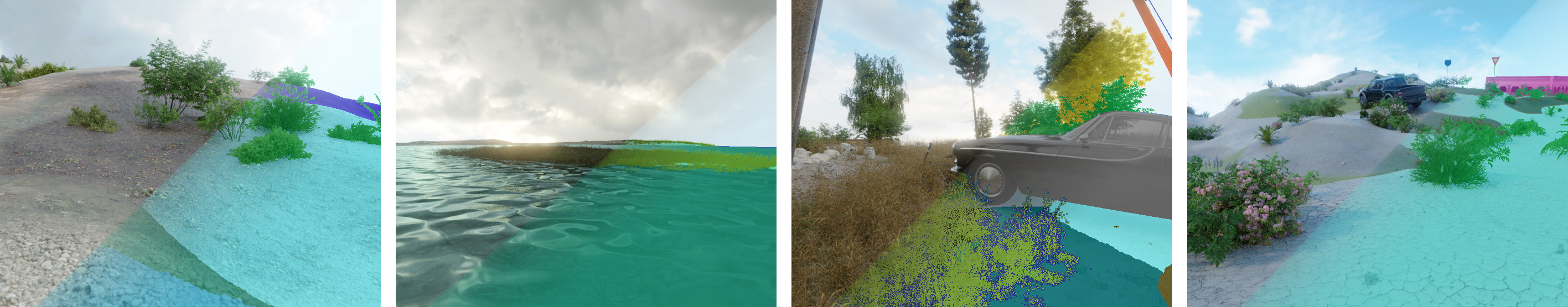}
\vspace{-1.5em}
\caption{Sample images from RUGDSynth overlaid with the corresponding annotations. Terrain regions are randomly colored in shades of cyan.}
\label{fig:rugdsynth01}
\end{figure*}

To train the class-agnostic component of our method, we require a large and diverse set of terrain samples.
In addition to covering a wide range of ground types, the dataset must include the mission-relevant, class-specific categories.
Since collecting such data in the real world is costly and time-consuming, we rely on synthetic data to efficiently meet the requirements.
Therefore, we extend OAISYS~\cite{muller_photorealistic_2021}, an open-source simulator pipeline based on Blender, that enables the creation of outdoor environments.
It relies on material-based semantics, ideally suitable for our use-case to simulate different terrains and their corresponding annotations.
Additionally, OAISYS provides the \textit{MeshParticleScatter} option, which enables to scatter meshes using the internal particle system of Blender. 
This is particularly useful for simulating large environments with numerous objects (\eg rocks, trees).
On the other hand, this concept is less suitable when simulating a smaller amount of objects (\eg cars, rovers) that have to be place more deliberately.
To this end, we extend the simulator with a physics-based placement module called \textit{MeshPhysicsScatter}. 
Objects can be spawned at a predefined height, where they fall under physical simulation onto a specified surface. 
Spawning locations can also be defined relative to the camera.
Given the large simulated environments, this gives a higher chance that the objects are visible in the camera frame.
Additionally, objects can be optionally respawned within a batch.
Originally, each object had to be listed in the config file, which is cumbersome, dealing with many objects.
We now allow a directory of assets, from which objects are randomly spawned.
We also extended the modules with a probabilistic activation parameter, controlling whether a module runs in a given batch. 
This prevents all assets from spawning simultaneously, increasing scene variability while keeping computation manageable.

\begin{table}[!t]
\caption{Used 3D Assets for RUGDSynth (total: 547)}
\vspace{-1em}
\begin{tabular}{lccc}
 \toprule
 3D Asset Type & No. Assets & Scatter Type & Activation Prob. [\%]\\
 \midrule
 Grass & 210 & Particle & 40.0 \\
 Tree & 96 & Particle & 60.0 \\
 Pole & 6 & Physical & 40.0 \\
 Vehicle & 41 & Physical & 99.2 \\
 Generic-Objects & 10 & Physical & 40.0 \\
 Building & 12 & Physical & 99.9 \\
 Log & 40 & Physical & 40.0 \\
 Bicycle & 5 & Physical & 30.0 \\
 Person & 26 & Physical & 40.0 \\
 Fence & 10 & Physical & 30.0 \\
 Bush & 23 & Particle & 70.0 \\
 Traffic Sign & 25 & Physical & 30.0 \\
 Rock & 41 & Particle & 27.0 \\
 Picnic-Table & 2 & Physical & 30.0 \\
 \bottomrule
\end{tabular}
\label{tab:assets}
\vspace{-1.3em}
\end{table}

OAISYS is further enhanced with a custom material module, \textit{MaterialCSCATerrain}, to simulate diverse ground surfaces. 
This module randomly selects textures from a predefined set and supports three configuration modes: class-specific assignment, class-agnostic assignment, or a mixture of both.
For each set, a minimum and maximum number of textures can be specified. 
The maximum number is particularly important to ensure that the number of terrain variations does not exceed the downstream network's capabilities. 
Additionally, this constraint helps maintain GPU memory consumption within reasonable limits. 
Furthermore, the developed module is not only applying selected textures using the default scales but is additionally randomizing the sizes of each texture to further increase visual variability.

Based on the OAISYS extensions, we created a synthetic version of RUGD~\cite{wigness_rugd_2019}, called \textit{RUGDSynth}.
The dataset features diverse ground types as well as the class-specific categories found in RUGD. 
Trees, grass, and rocks are distributed across the scene using a particle system, with objects (e.g., vehicles, logs) spawned at the initial sensor location. Since the sensor is not moved far from its initial location, respawning within a batch is unnecessary, reducing computational overhead. \cref{tab:assets} gives an overview over the used mesh assets.
The sky is simulated using 68 randomly selected HDR images with randomized emission intensity. 
In total, 204 unique ground textures are used.

RUGDSynth contains 42,672 samples across more than 4,500 worlds (batches), each with roughly 10 samples. 
The dataset is split into 42,672 training, 1,000 validation, and 1,924 test samples. 
\Cref{fig:rugdsynth01} shows example scenes.

\section{Experiments}
In this section, we present different experiments to demonstrate our method.
Using RUGD~\cite{wigness_rugd_2019} as a benchmark, all ground surfaces are treated as class-agnostic, while mission-relevant, platform- and context-independent categories are labeled as class-specific.
Consequently, terrains such as sand, gravel, and mud, including their sub-classes, are assigned to the class-agnostic set, while trees, vegetation, vehicles, and buildings are treated as class-specific.
In principle, any separation into different subsets should be possible and is independent of the method presented here.

\subsection{Evaluation Metric}

\begin{figure}[t!]
\centering
\vspace{0.5em}
\includegraphics[width=\linewidth]{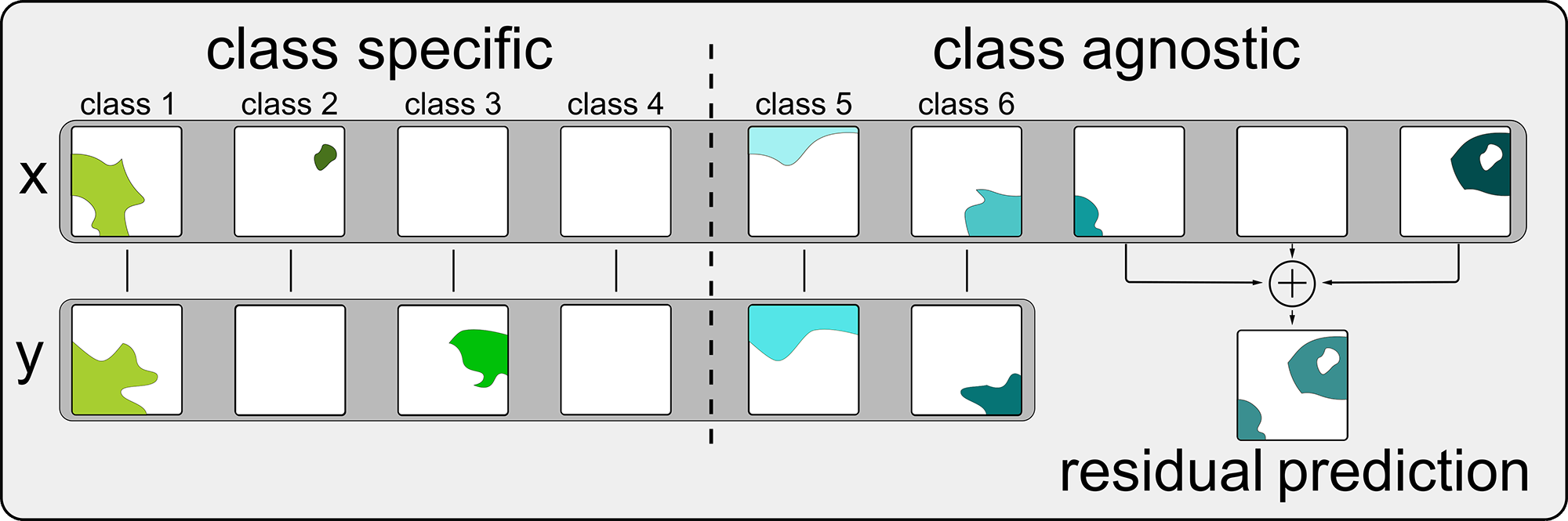}
\vspace{-1.5em}
\caption{Illustration of the residual prediction calculation. All region proposal masks that are not matched to a ground-truth mask are summed to produce the residual prediction mask. In the ideal case, this mask is empty.}
\vspace{-1em}
\label{fig:eval_figure}
\end{figure}

In semantic segmentation, the number of classes in both the annotations and the predictions is predefined and consistent.
This assumption is not given for the class-agnostic task.
Although the number of annotations is fixed, the number of potential region predictions is arbitrary and determined by the method.
Relying solely on the standard \gls{IoU} values would fail to account for errors arising from predicting more class-agnostic regions than exist.
Consequently, the \gls{IoU} as standard semantic segmentation metric is insufficient for our case, therefore, we introduce an additional metric.
Specifically, we aggregate all residual class-agnostic prediction masks that were not matched with any ground truth annotation mask into a single mask.
Ideally, the consquent residual mask should contain no entries, whereas in the worst case, it has entries at every pixel.
Based on the residual mask, we calculate the recall value, comparing it to a ground truth mask where all entries are set to one.
The resulting \gls{resR} value quantifies the ratio of predicted to actual class-agnostic terrain regions in the scene.
Larger values indicate an increasing surplus of predicted regions.
\cref{fig:eval_figure} illustrates the metric calculation.

\subsection{Evaluation Data}
For the evaluation we use two datasets, which are briefly described below.
Each dataset has unique characteristics.
The first dataset, RUGD, is well-established and has also been used for training our method.
To showcase transferability, we introduce a second dataset: EXTerra - Plan\textbf{e}tary E\textbf{x}ploration \textbf{Terr}ain Dataset.
Importantly, no data from the latter dataset is used for training the underlying method.
Moreover, EXTerra includes additional ground types that are not covered by the RUGD dataset.

\subsubsection{RUGD Dataset}
The RUGD dataset features data collected by a mobile robot traversing in a variety of unstructured and semi-structured outdoor environments. 
Since it provides many different ground types it is ideally suitable as benchmark for our evaluation. 
It provides 25 semantic classes, from which one class is "void" and two of the classes ("bridge" and "bicycle") are not part of the test set, therefore ignored. 
We split the categories into a set of specific classes and class-agnostic terrains, listed in \cref{table:cs_results_rugd} and \cref{table:ca_results_rugd}, respectively.
Consequently, there are 16 specific classes and 8 class-agnostic terrains.
Our network features the therefore also 16 semantic classes, and provides 20 possible class-agnostic prediction slots.
If not mentioned differently, we train \model~with the training set of RUGD and RUGDSynth. 

\subsubsection{EXTerra Dataset}
\begin{figure*}[h]
\centering
\includegraphics[width=\textwidth]{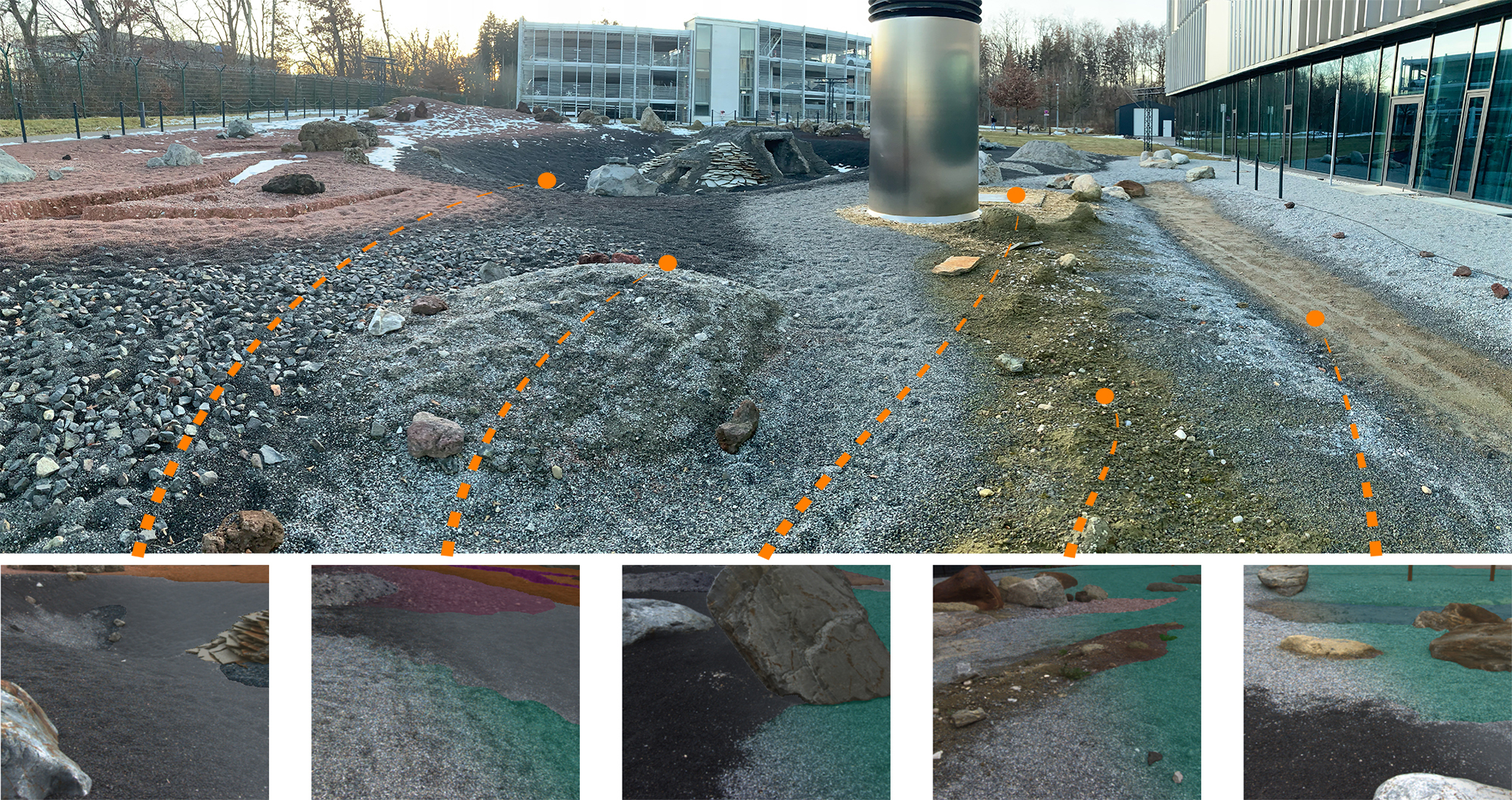}
\vspace{-1.3em}
\caption{Image of the planetary exploration outdoor laboratory and example samples from the EXTerra dataset with partially overlaid annotations and the approximate capture locations indicated by orange lines.}
\vspace{-1.5em}
\label{fig:dlr_outdoor_01}
\end{figure*}
As an additional dataset to evaluate the domain-shift capacity, we collected and annotated data from a planetary exploration outdoor lab.
The data were acquired using a rover system.
The semantic categories are labeled according to the RUGD taxonomy, while the terrain regions are agnostically annoated without further semantic distinction.
These terrains are considered sub-classes of sand, mud, and gravel surfaces.
The dataset comprises 14 additional distinct terrain types and consists of 124 samples.
\Cref{fig:dlr_outdoor_01} gives an overview over the outdoor lab and examples from the EXTerra Dataset.

\subsection{Class-Specific Evaluation}
\begin{figure*}[ht]
\centering
\vspace{1em}
\resizebox{1\textwidth}{!}{ 
\begin{tabular}{c ccccccc}   
    \rotatebox{90}{\parbox{3.5cm}{\centering \textbf{Inputs}}} &
    \includegraphics[width=0.2\textwidth]{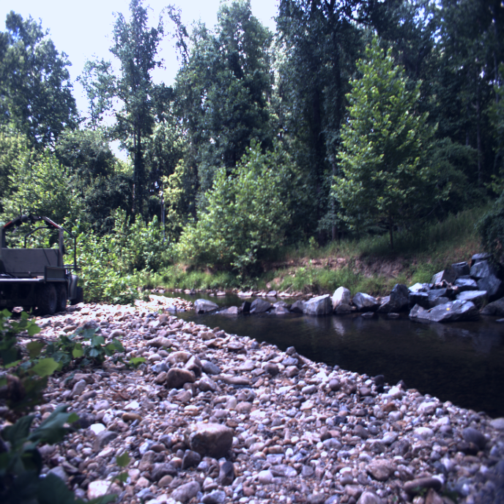} &
    \includegraphics[width=0.2\textwidth]{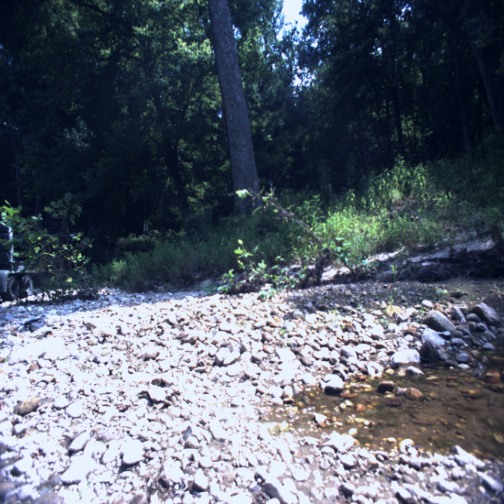} &
    \includegraphics[width=0.2\textwidth]{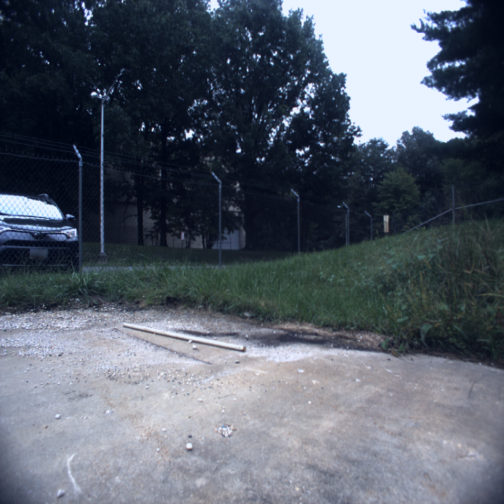} &
    \includegraphics[width=0.2\textwidth]{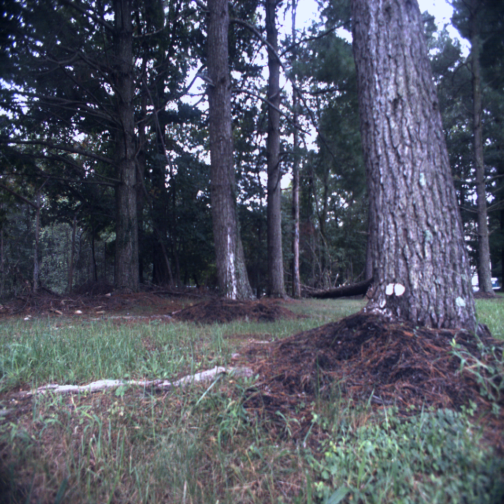} &
    \includegraphics[width=0.2\textwidth]{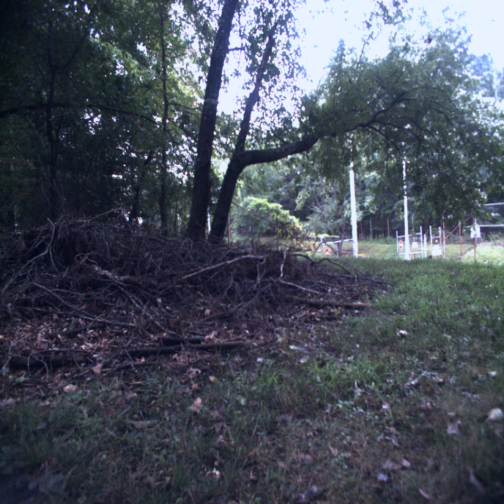} &
    \includegraphics[width=0.2\textwidth]{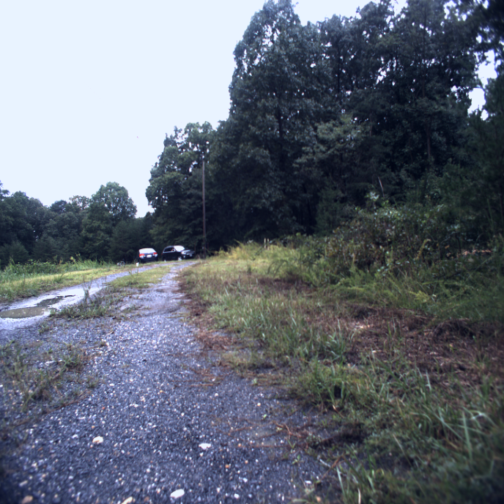} &
    \includegraphics[width=0.2\textwidth]{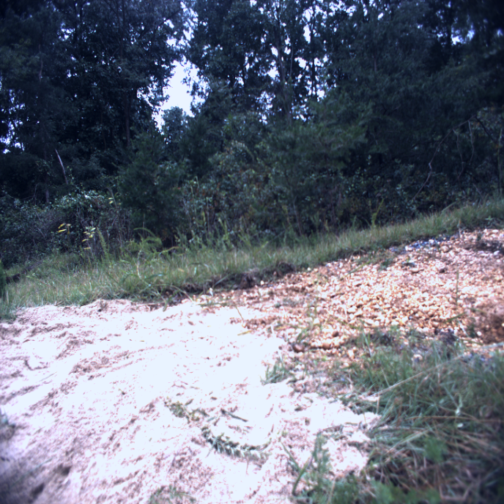} \\
    \rotatebox{90}{\parbox{3.5cm}{\centering \textbf{Targets}}} &
    \includegraphics[width=0.2\textwidth]{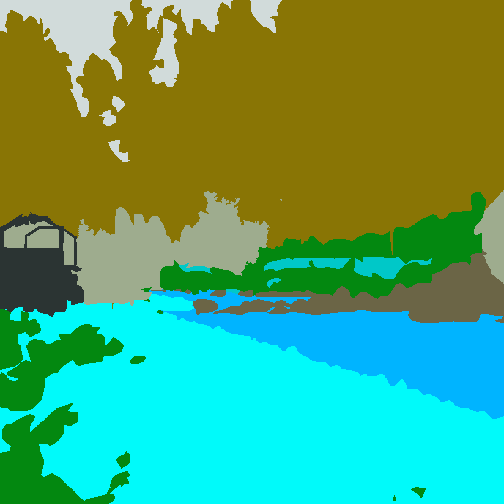} &
    \includegraphics[width=0.2\textwidth]{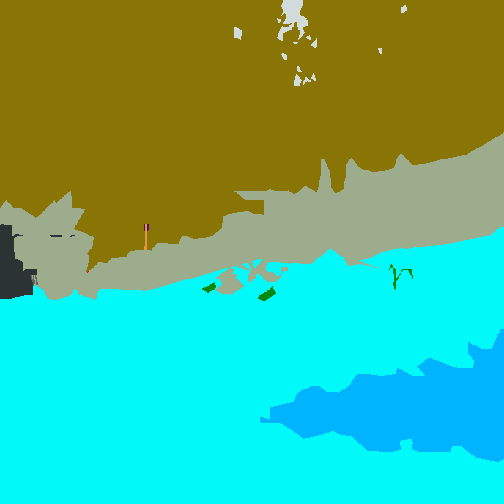} &
    \includegraphics[width=0.2\textwidth]{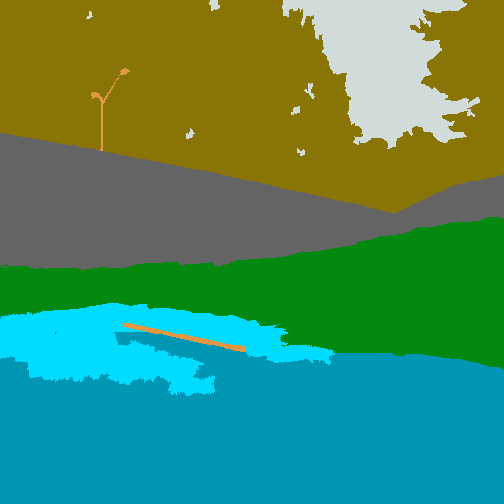} &
    \includegraphics[width=0.2\textwidth]{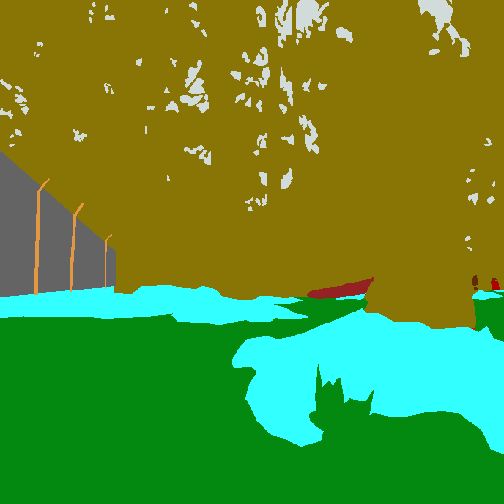} &
    \includegraphics[width=0.2\textwidth]{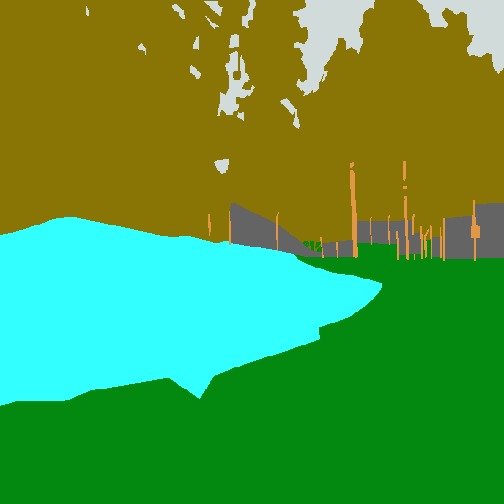} &
    \includegraphics[width=0.2\textwidth]{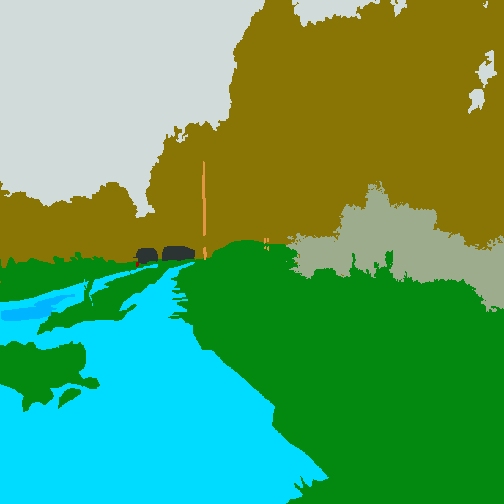} &
    \includegraphics[width=0.2\textwidth]{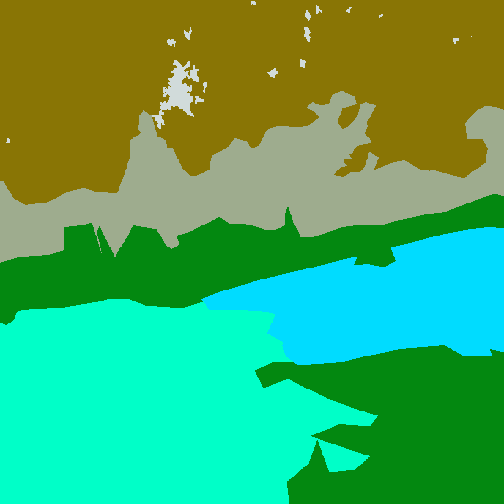} \\
    \rotatebox{90}{\parbox{3.5cm}{\centering \textbf{\model}}} &
    \includegraphics[width=0.2\textwidth]{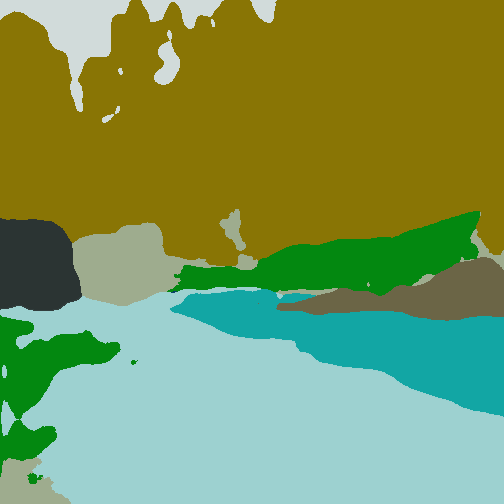} &
    \includegraphics[width=0.2\textwidth]{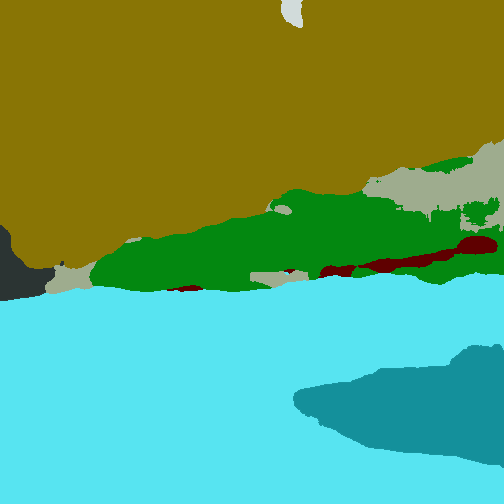} &
    \includegraphics[width=0.2\textwidth]{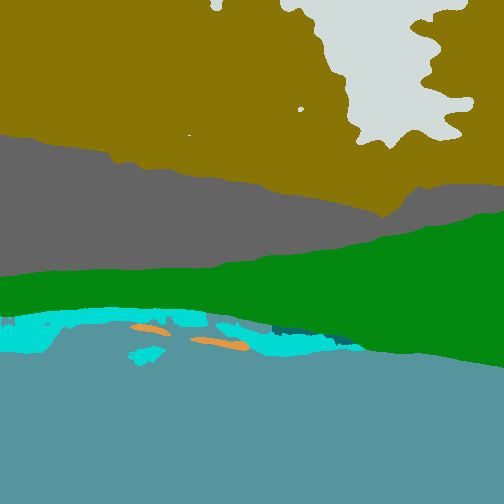} &
    \includegraphics[width=0.2\textwidth]{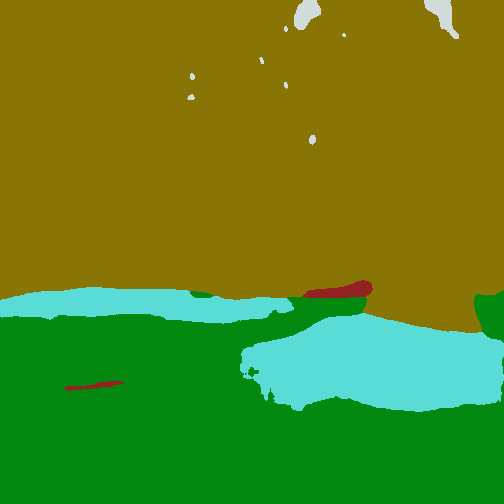} &
    \includegraphics[width=0.2\textwidth]{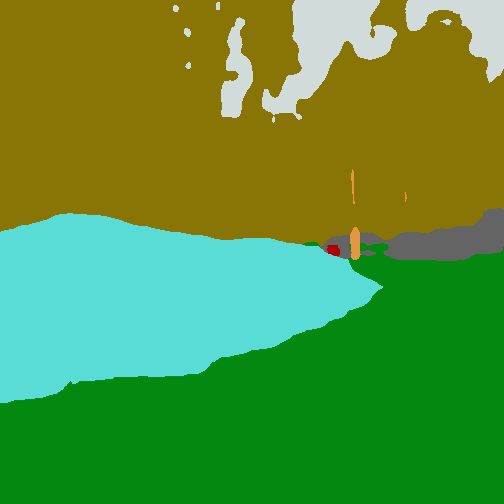} &
    \includegraphics[width=0.2\textwidth]{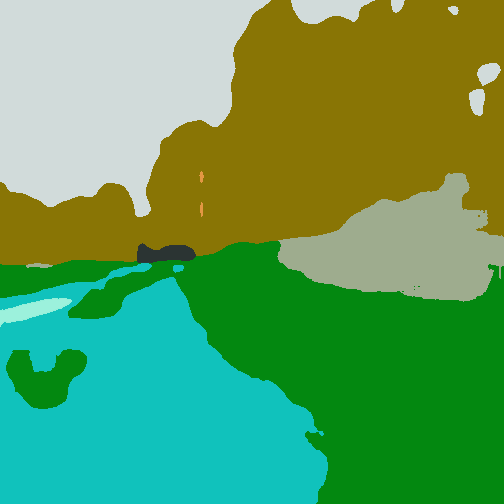} &
    \includegraphics[width=0.2\textwidth]{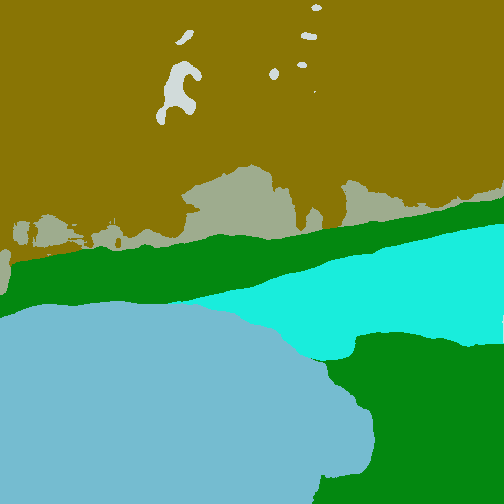} \\
    \rotatebox{90}{\parbox{3.5cm}{\centering \textbf{SAM}}} &
    \includegraphics[width=0.2\textwidth]{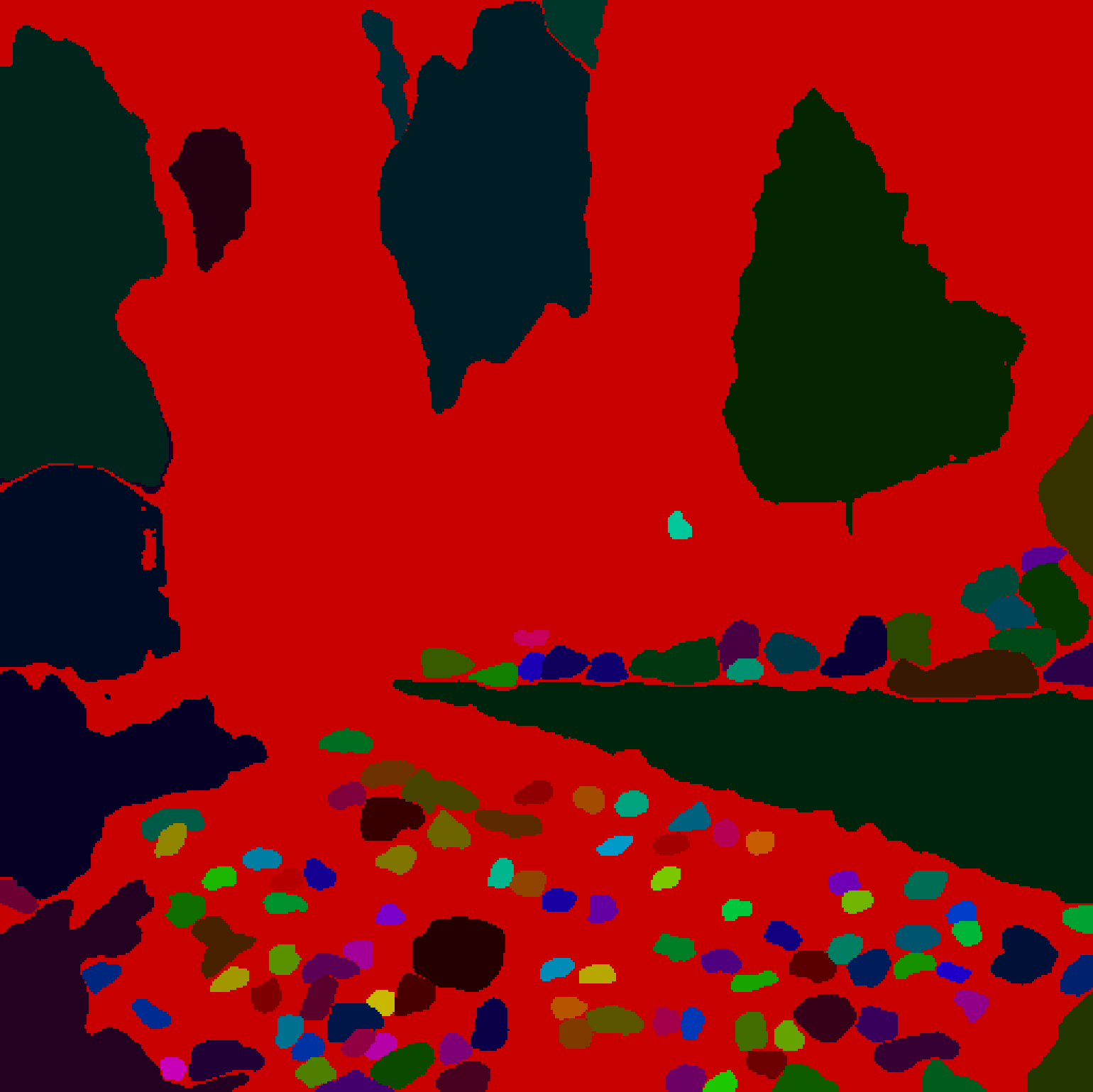} &
    \includegraphics[width=0.2\textwidth]{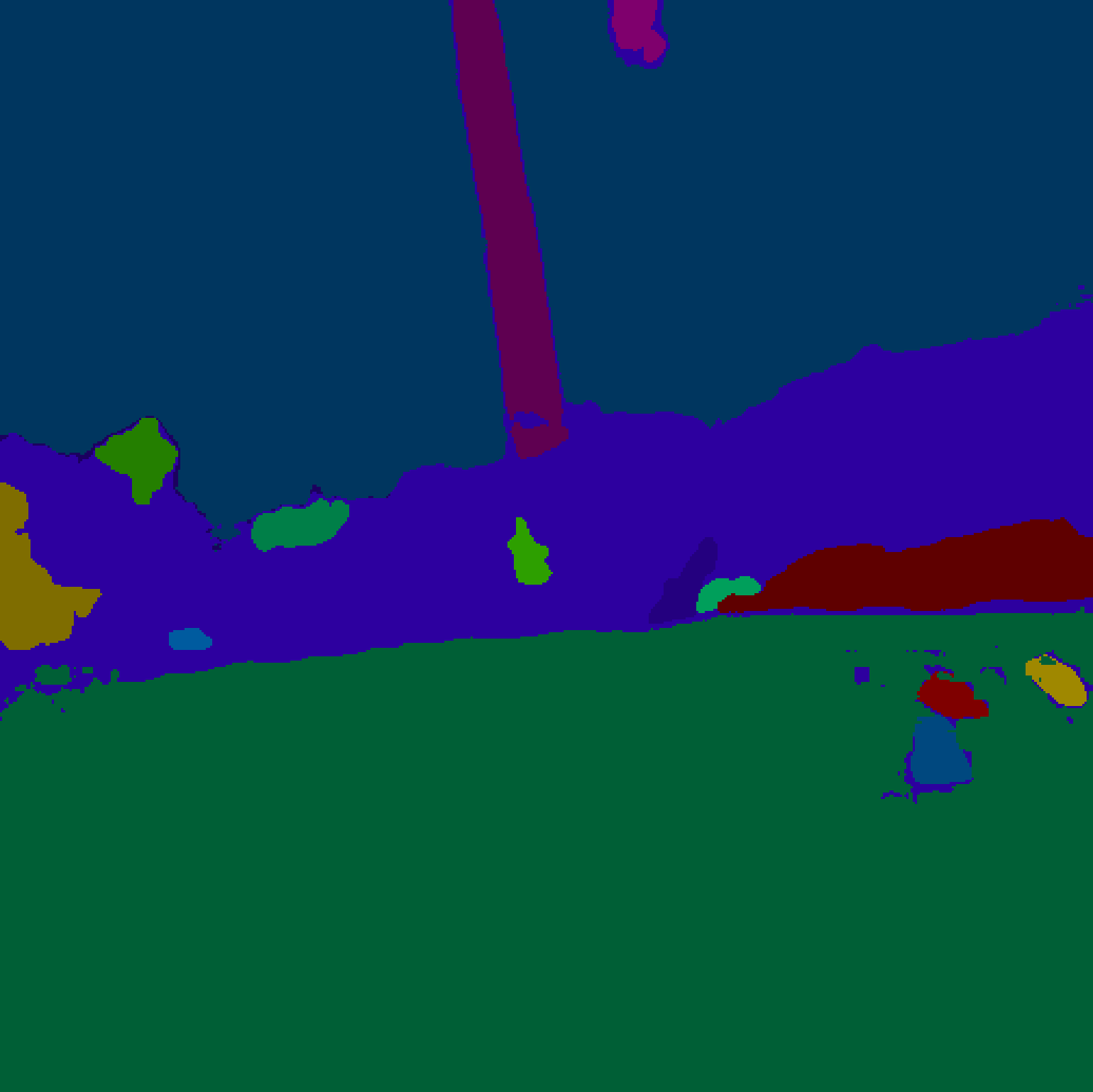} &
    \includegraphics[width=0.2\textwidth]{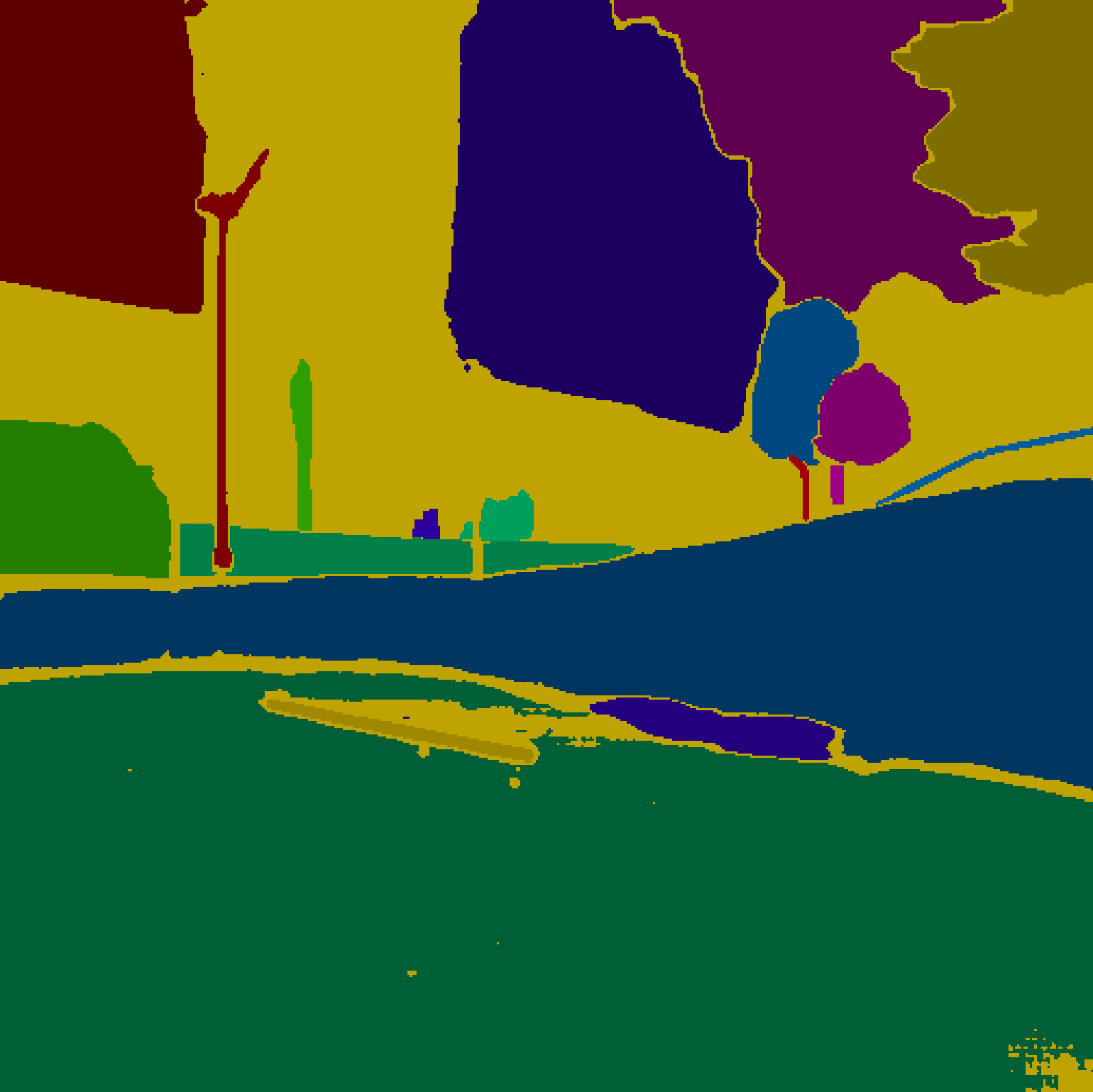} &
    \includegraphics[width=0.2\textwidth]{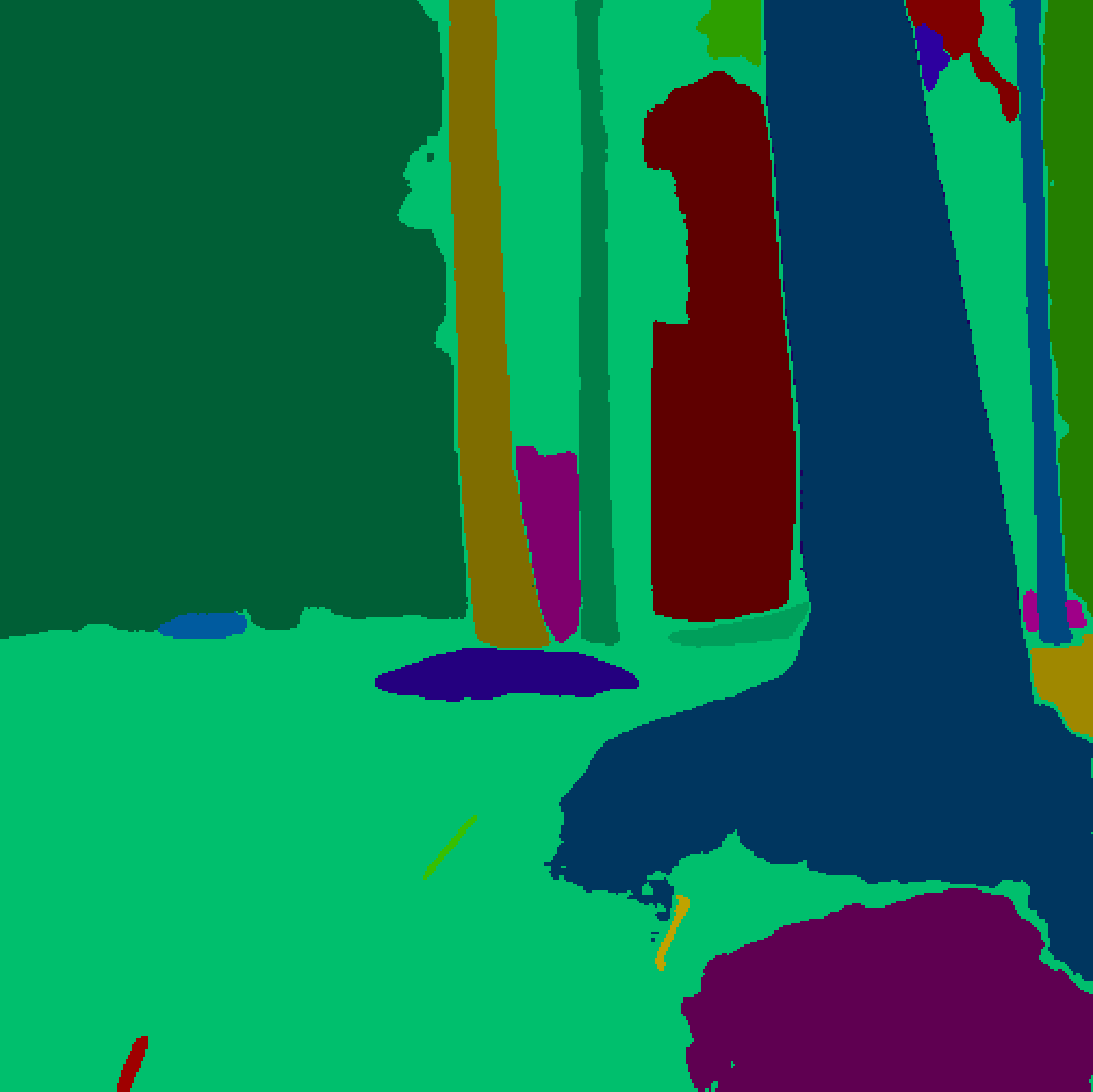} &
    \includegraphics[width=0.2\textwidth]{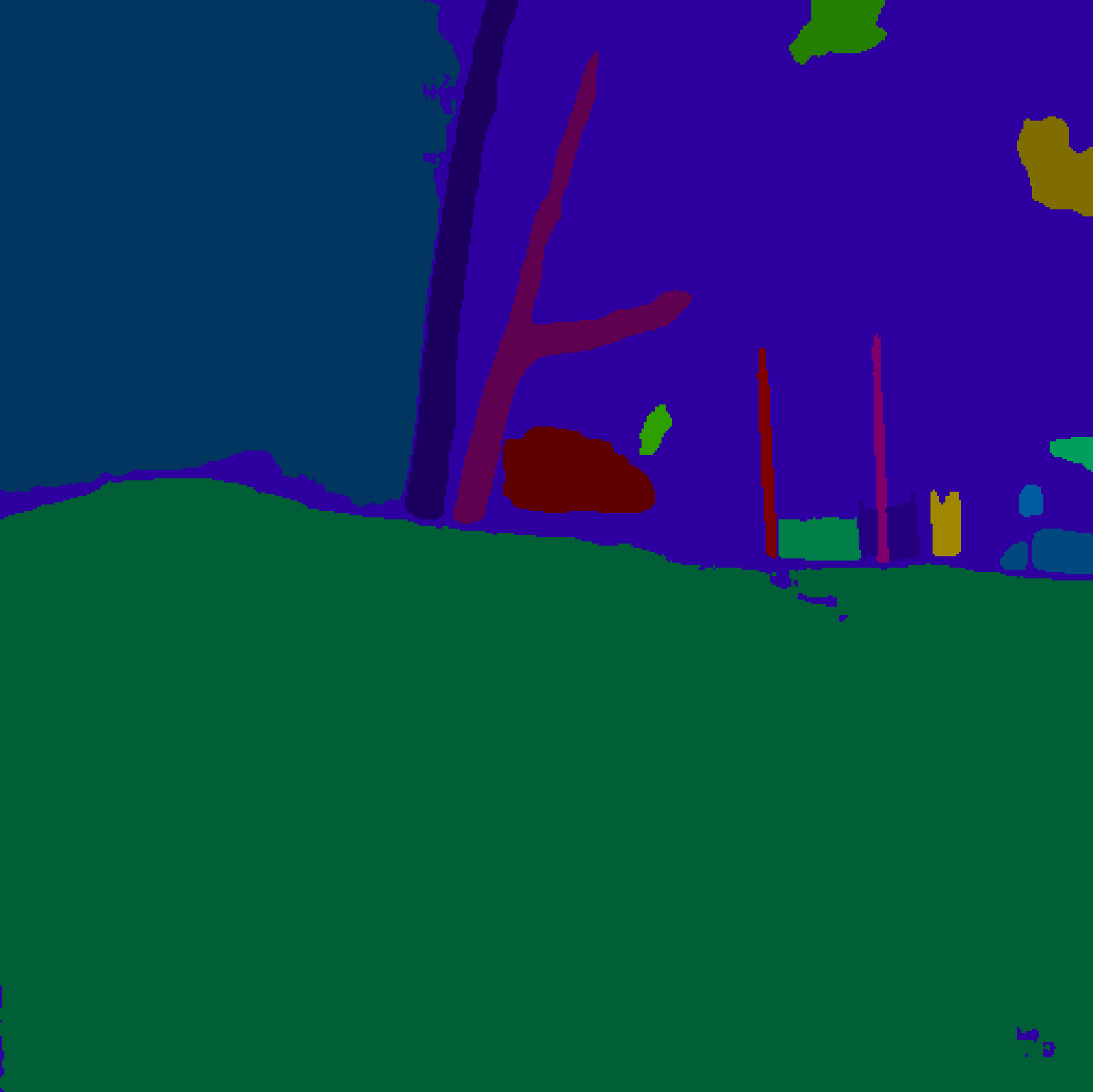} &
    \includegraphics[width=0.2\textwidth]{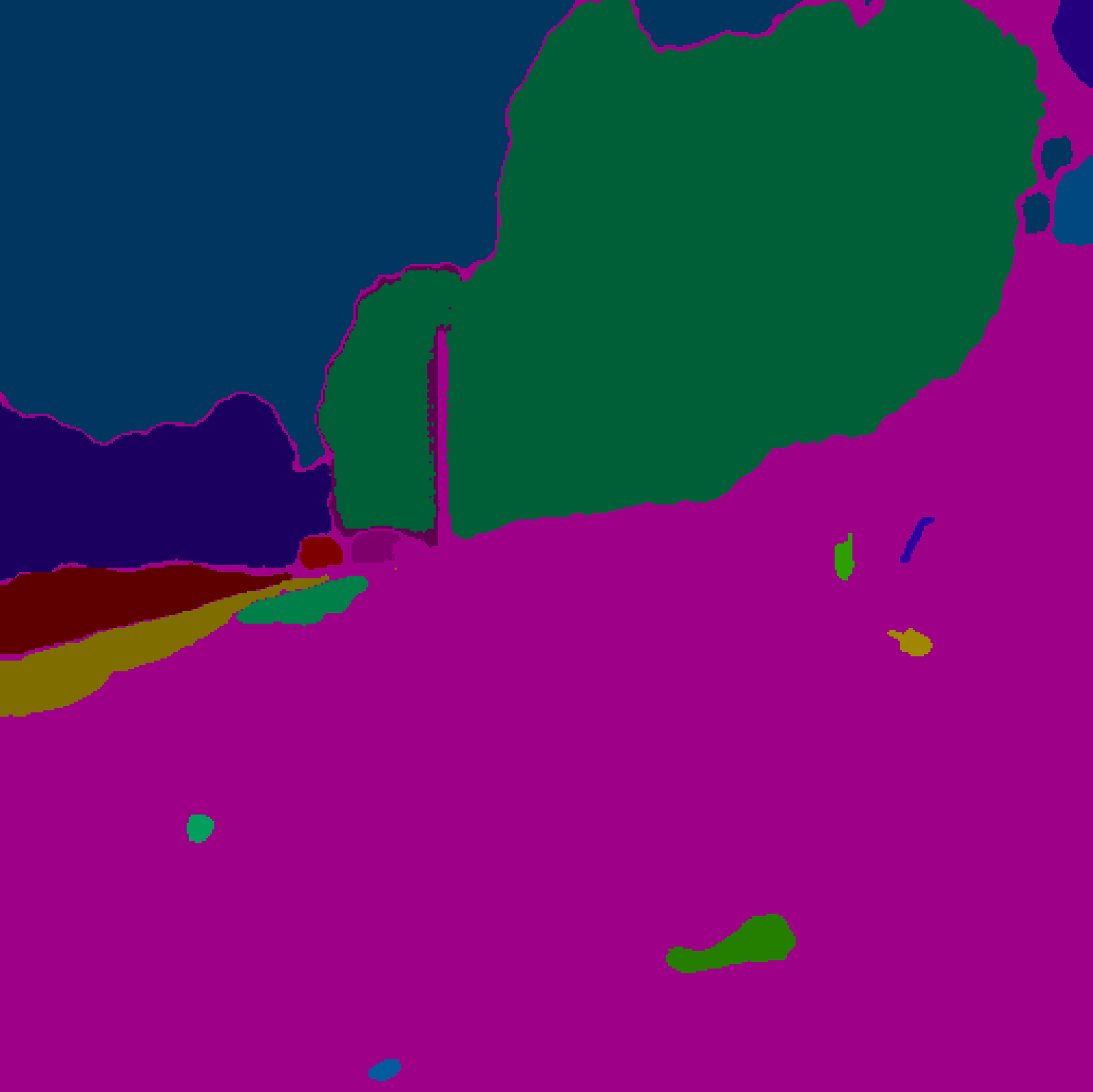} &
    \includegraphics[width=0.2\textwidth]{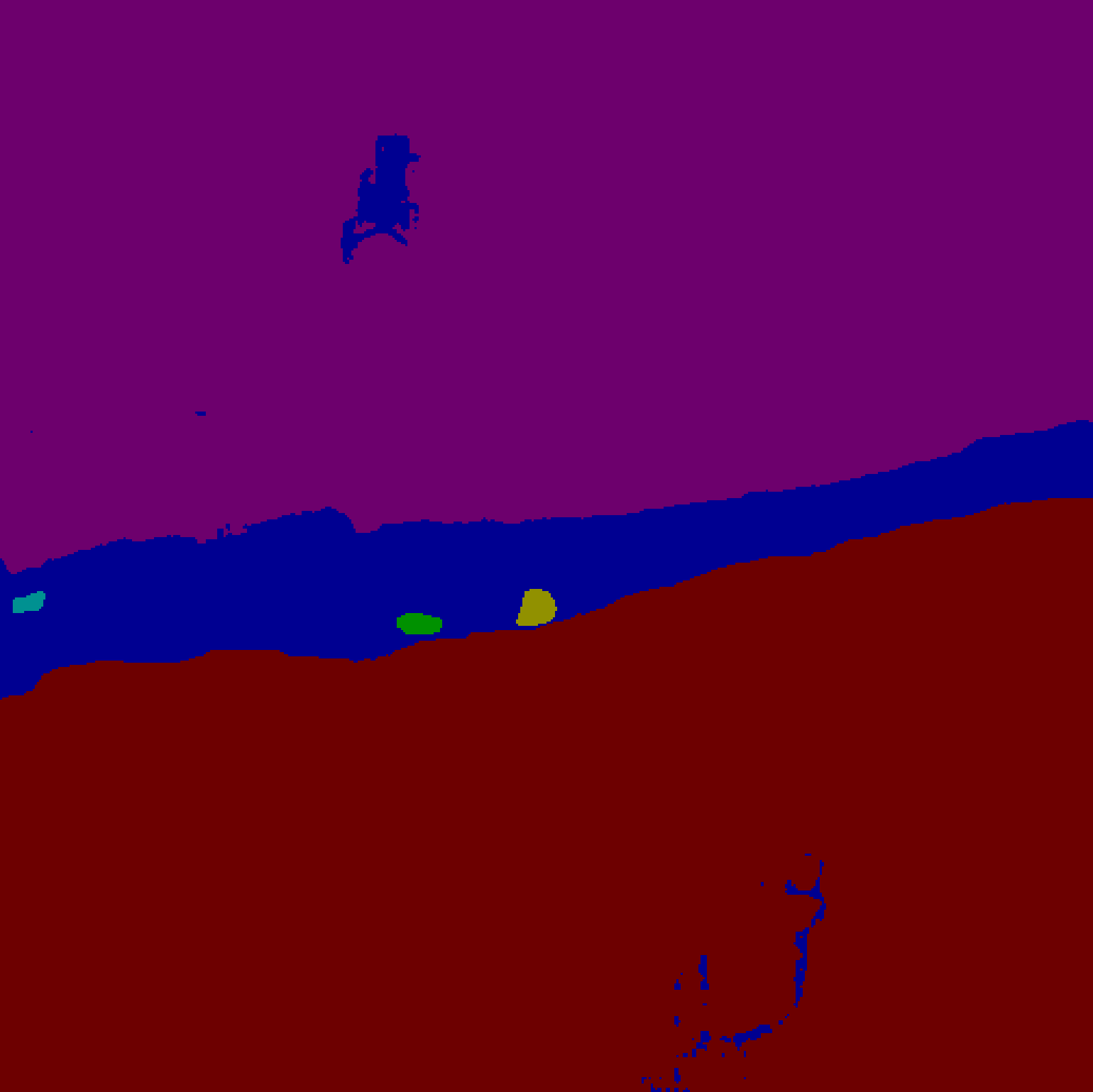} \\
\end{tabular}
}

\caption{Selected qualitative results for the RUGD dataset. Class-agnostic regions are colored in shades of cyan. Note that the cyan colors used for the class-agnostic predictions are randomly assigned for each sample and do not necessarily match the ground-truth colors. For visualization clarity, the region predictions of SAM are randomly selected.}
\label{fig:predictions_rugd}
\end{figure*}

In this evaluation we focus on the class-specific segmentation capabilities of our \model.
\Cref{table:cs_results_rugd} lists the class-specific IoU values and the mean IoU (mIoU) over all class-specific classes. 
We compare our method against the m1 model~\cite{wigness_rugd_2019}. 
Furthermore, we report CRLNet~\cite{li_crlnet_2026}, which to the best of our knowledge is the current best performing segmentation network on the RUGD dataset. 
Additionally, we list \gls{SAM}~\cite{kirillov_segment_2023} as class-agnostic region proposal generator.
Our model achieves the highest mIoU. 
CRLNet achieves high values in many specific classes, however lacks to segment more rare classes like "person" and "sign".
Qualitative results can be seen in~\cref{fig:predictions_rugd}.

Quantitative results on the EXTerra Dataset are provided in~\cref{table:synth_real_ca_results_dlr}, while qualitative results are found in~\cref{fig:predictions_dlr}.
\gls{SAM} is performing slightly better in the class-specific case.
We hypothesise that most appearance of such classes are in the far field of the robot.
Since SAM is having a uniformly grid of point prompts, it does not differentiate too much between fore- and background.
Notice that SAM has troubles segmenting correct class-specific regions on which it was not explicitly trained, as can be seen with the "fence" class in column three of~\cref{fig:predictions_rugd}.

\begin{table}[b!]
\centering
\vspace{-1.2em}
\caption{Class-Specific IoU Evaluation of RUGD}
\vspace{-1.5em}
\resizebox{\linewidth}{!}{
\begin{tabular}[t]{lS[table-format=2.2]S[table-format=2.2]S[table-format=2.2]S[table-format=2.2]}
\toprule
class&\multicolumn{1}{c}{m1\cite{wigness_rugd_2019}}&\multicolumn{1}{c}{CRLNet\cite{li_crlnet_2026}}& \multicolumn{1}{c}{SAM\cite{kirillov_segment_2023}} &\multicolumn{1}{c}{\model~(ours)}\\
\midrule
grass &73.31& \textbf{86.10} & 73.27 & 78.49\\
tree &79.76& \textbf{92.00} & 59.31 & 84.12\\
pole &15.57& \textbf{34.90} & 14.29 & 22.13\\
sky &\textbf{79.16}& 73.70 & 54.3 & 78.13\\
vehicle &54.47 & \textbf{75.50} & 50.44 & 75.29\\
generic-object &1.10& \textbf{44.90} & 3.89 & 11.66\\
building &73.44& 41.60 & 49.04 & \textbf{78.81}\\
log &43.13 & 40.60 & 40.56 & \textbf{57.82}\\
person &0.00& 0.00 & 7.56 & \textbf{24.92}\\
fence &52.53& 32.30 & 21.70 & \textbf{63.20}\\
bush &20.93& 30.20 & \textbf{32.81} & 28.93\\
sign &7.70 & 0.00 & 9.74 & \textbf{29.80}\\
rock &12.51& \textbf{55.30} & 20.66 & 32.46\\
picnic-table &55.96& \textbf{75.20} & 36.93 & 59.36\\
\midrule
mIoU&40.68&48.74&33.89&\textbf{51.80}\\
\bottomrule
\end{tabular}
}
\label{table:cs_results_rugd}
\end{table}

\subsection{Class-Agnostic Evaluation}
Next we are looking at the class-agnostic terrain segmentation part of the network. Results for the RUGD dataset are shown in~\cref{table:ca_results_rugd}. \model~has the highest IoU value for all classes except of one, resulting in the highest mIoU.
Precision, Recall, and Residual Recall are additionally reported for SAM and \model~in~\cref{table:ca_results_rugd}.

Qualitative results are illustrated in~\cref{fig:predictions_rugd}.
Notice that SAM tends to oversegment the image in the foreground, when more details of a terrain appear and also struggles to segment relevant terrains.
In contrast, \model~is able to distinguish the terrains in these examples.

\begin{table}[!b]
\centering
\vspace{-1.5em}
\caption{Class-Agnostic Evaluation of RUGD}
\vspace{-1.5em}
\resizebox{\linewidth}{!}{
\begin{tabular}[t]{lS[table-format=2.2]S[table-format=2.2]S[table-format=2.2]S[table-format=2.2]}
\toprule
class&\multicolumn{1}{c}{m1\cite{wigness_rugd_2019}}&\multicolumn{1}{c}{CRLNet\cite{li_crlnet_2026}}& \multicolumn{1}{c}{SAM\cite{kirillov_segment_2023}} &\multicolumn{1}{c}{\model~(ours)}\\
\midrule
dirt & 0.48 & 0.0 & 16.99 & \textbf{34.83}\\
sand & 40.17 & 0.0 & 26.81 & \textbf{60.26}\\
water & 5.30 & 55.4 & 30.42 & \textbf{82.04}\\
asphalt & 12.47 & \textbf{94.00} & 17.79 & 29.0\\
gravel &33.94 & 81.1 & 56.99 & \textbf{86.19}\\
mulch & 49.71& 82.9 & 70.09 & \textbf{84.15}\\
rock-bed &9.77 & 0.0 & 45.97 & \textbf{86.17}\\
concrete & 84.71& 13.9 & 12.44 & \textbf{89.44}\\
\midrule
mIoU&29.57&40.91&34.69&\textbf{69.01}\\
\midrule
mPre & \emph{n/a} & \emph{n/a} & 48.96 & \textbf{75.20}\\
mRec & \emph{n/a} & \emph{n/a} & 38.94 & \textbf{54.84} \\
resR $\downarrow$ & \emph{n/a} & \emph{n/a} & 17.67 & \textbf{0.28} \\
\bottomrule
\end{tabular}
}
\label{table:ca_results_rugd}
\end{table}

Results for the EXTerra Dataset are listed in~\cref{table:synth_real_ca_results_dlr}.
\model~outperforms \gls{SAM} in the class-agnostic case by a high margin.
In \cref{fig:predictions_dlr} we can observe that SAM is preforming well when boundaries are visually clear.
If boundaries are fuzzy and intersect, like it is often the case for terrains, the mask predictions quality is decreasing as can be seen in the example of second column of \cref{fig:predictions_dlr}.

\subsection{Relevance of Synth Data Evaluation}
\begin{figure*}[ht]
\centering

\resizebox{1\textwidth}{!}{  
\begin{tabular}{c ccccccc}  
    \rotatebox{90}{\parbox{4.0cm}{\centering \textbf{Inputs}}} &
    \includegraphics[width=0.2\textwidth]{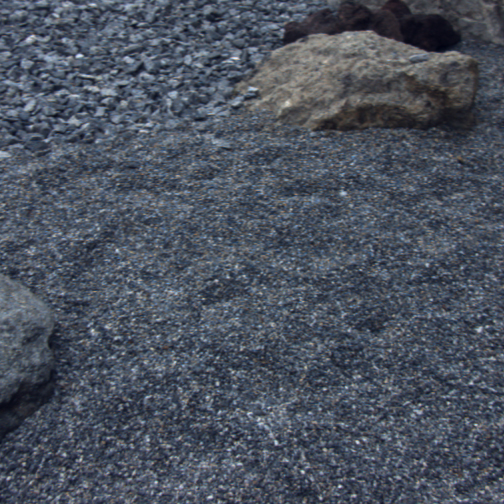} &
    \includegraphics[width=0.2\textwidth]{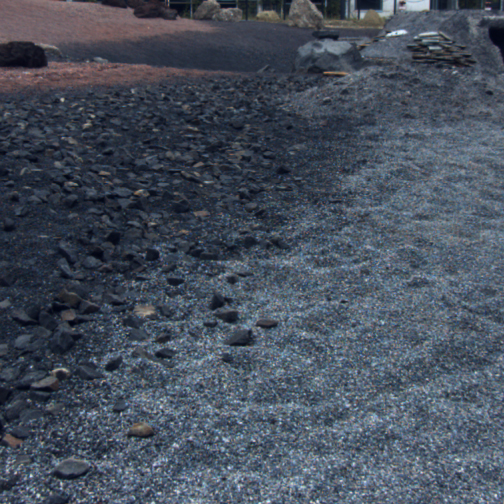} &
    \includegraphics[width=0.2\textwidth]{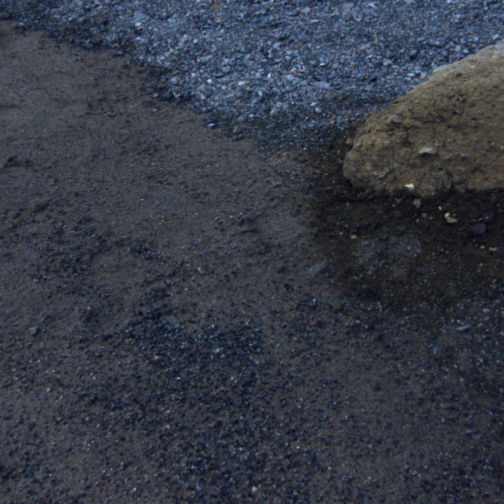} &
    \includegraphics[width=0.2\textwidth]{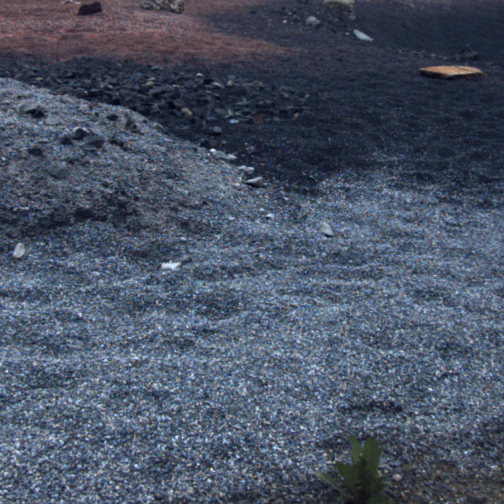} &
    \includegraphics[width=0.2\textwidth]{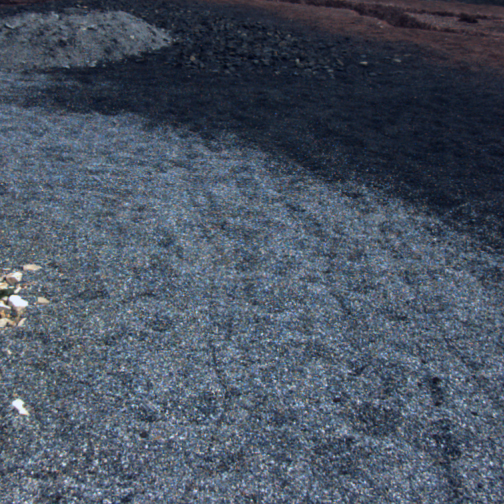} &
    \includegraphics[width=0.2\textwidth]{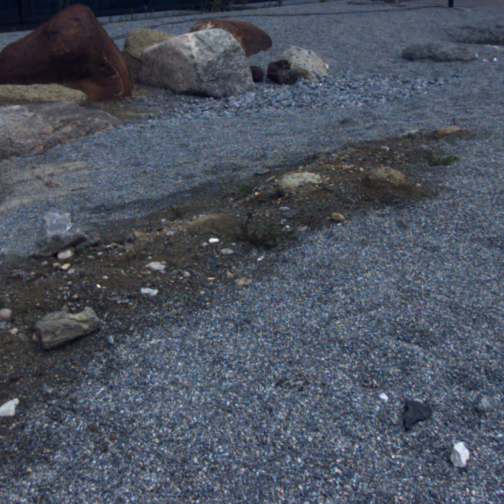} &
    \includegraphics[width=0.2\textwidth]{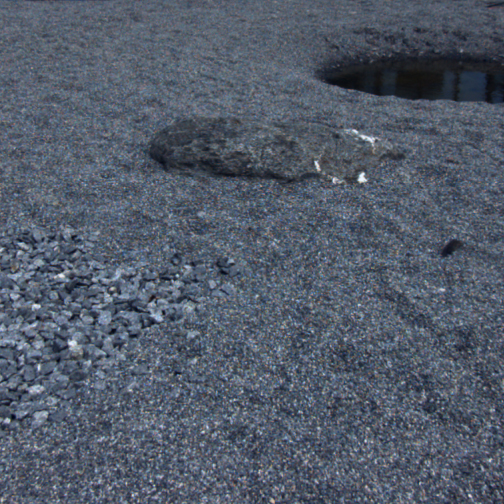} \\
    \rotatebox{90}{\parbox{4.0cm}{\centering \textbf{Targets}}} &
    \includegraphics[width=0.2\textwidth]{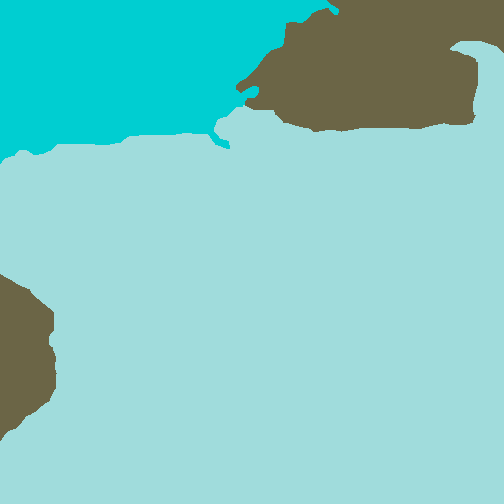} &
    \includegraphics[width=0.2\textwidth]{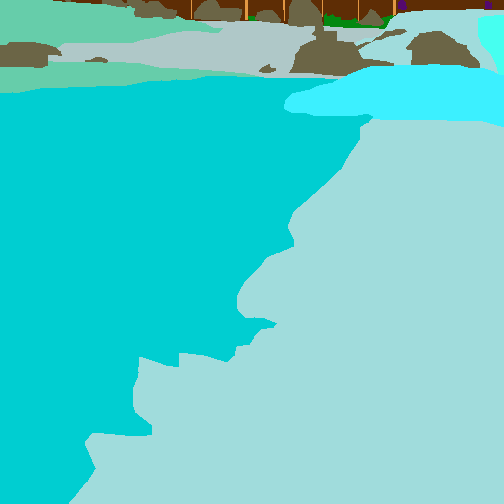} &
    \includegraphics[width=0.2\textwidth]{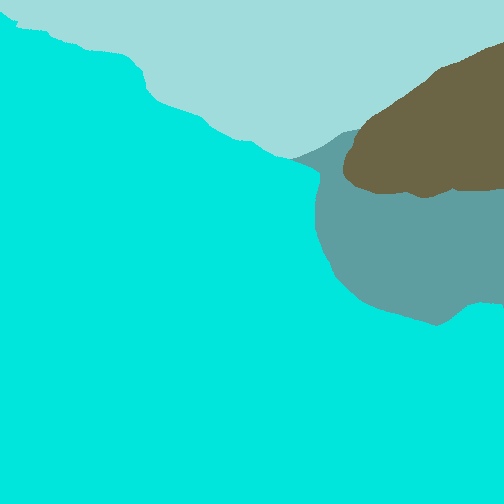} &
    \includegraphics[width=0.2\textwidth]{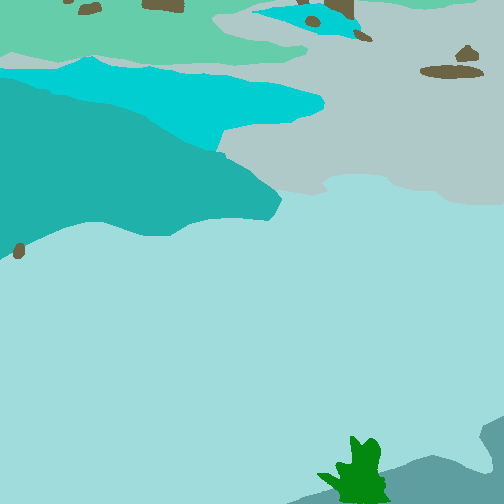} &
    \includegraphics[width=0.2\textwidth]{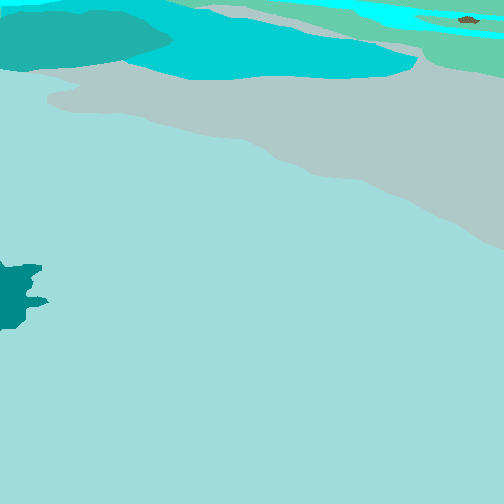} &
    \includegraphics[width=0.2\textwidth]{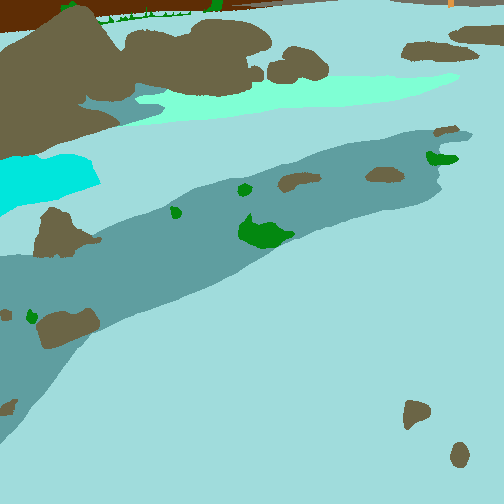} &
    \includegraphics[width=0.2\textwidth]{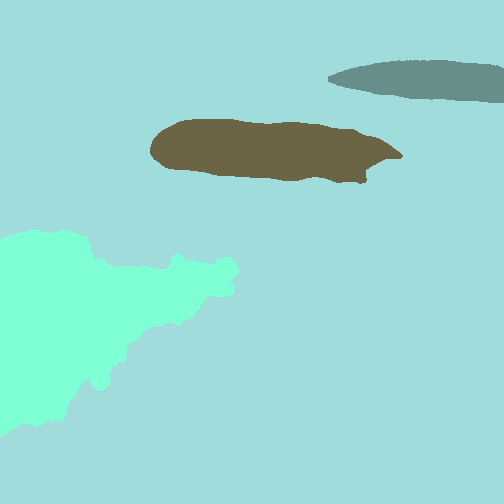} \\
    \rotatebox{90}{\parbox{4.0cm}{\centering \textbf{\model}}} &
    \includegraphics[width=0.2\textwidth]{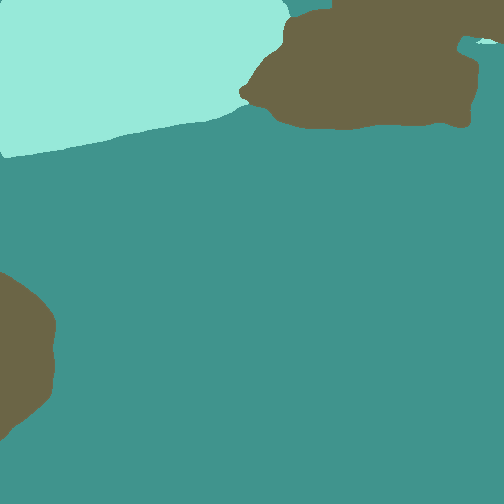} &
    \includegraphics[width=0.2\textwidth]{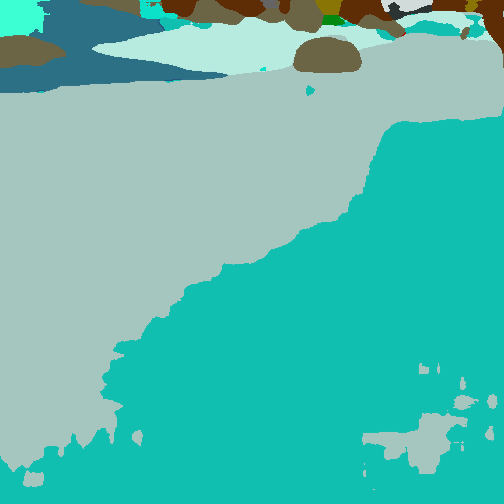} &
    \includegraphics[width=0.2\textwidth]{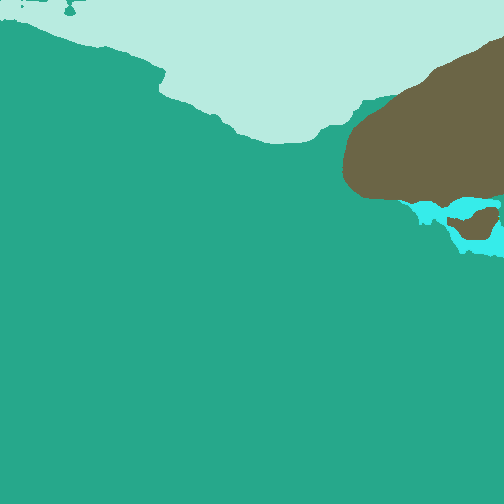} &
    \includegraphics[width=0.2\textwidth]{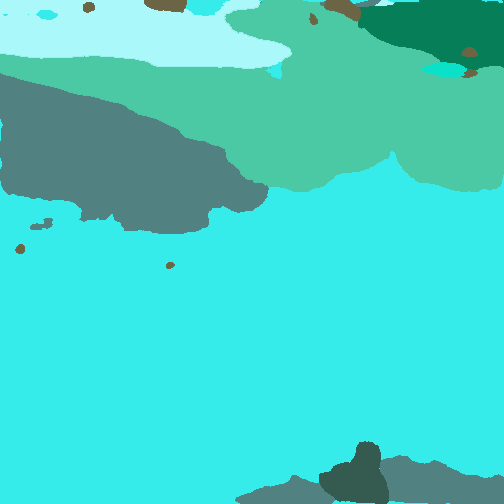} &
    \includegraphics[width=0.2\textwidth]{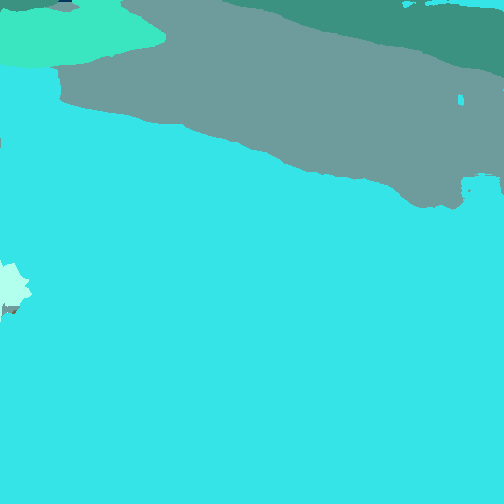} &
    \includegraphics[width=0.2\textwidth]{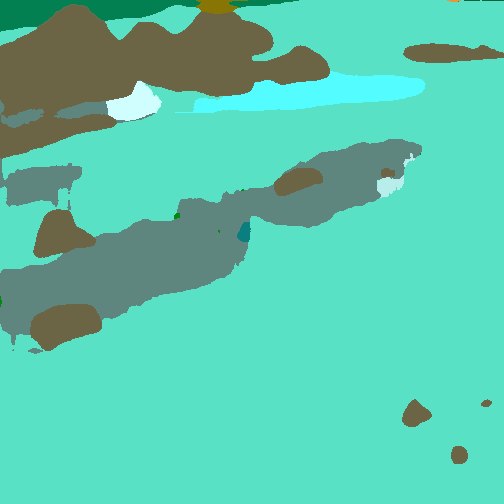} &
    \includegraphics[width=0.2\textwidth]{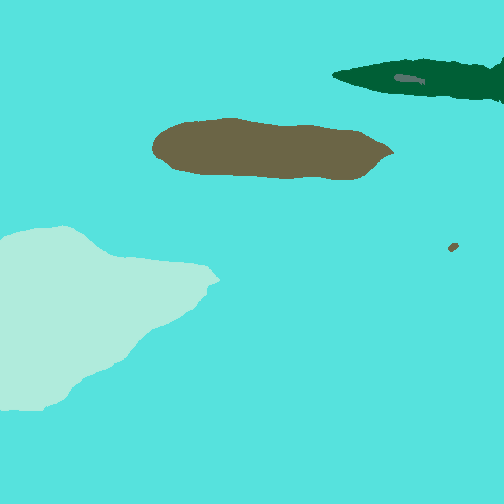} \\
    \rotatebox{90}{\parbox{4.0cm}{\centering \textbf{SAM}}} &
    \includegraphics[width=0.2\textwidth]{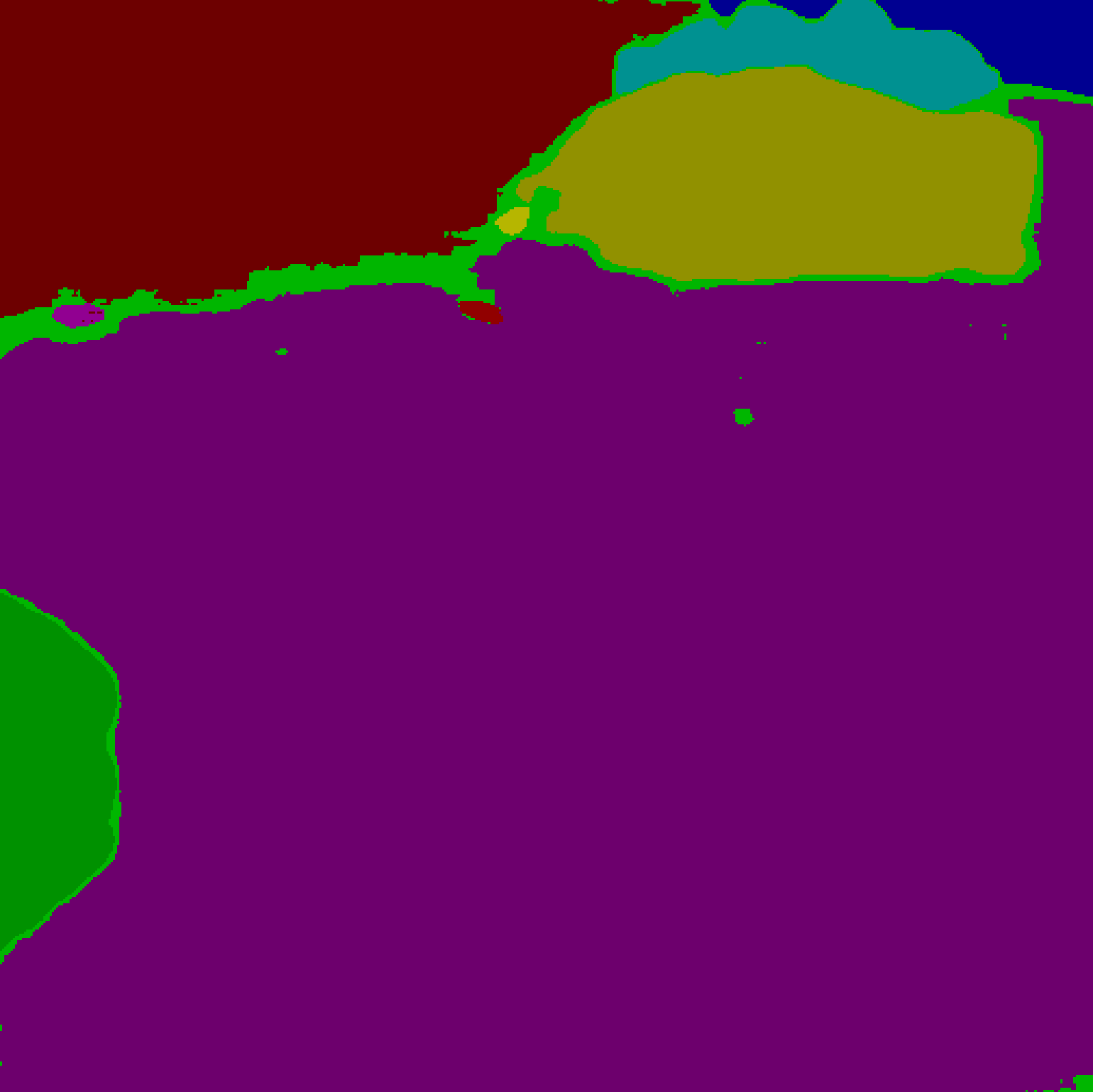} &
    \includegraphics[width=0.2\textwidth]{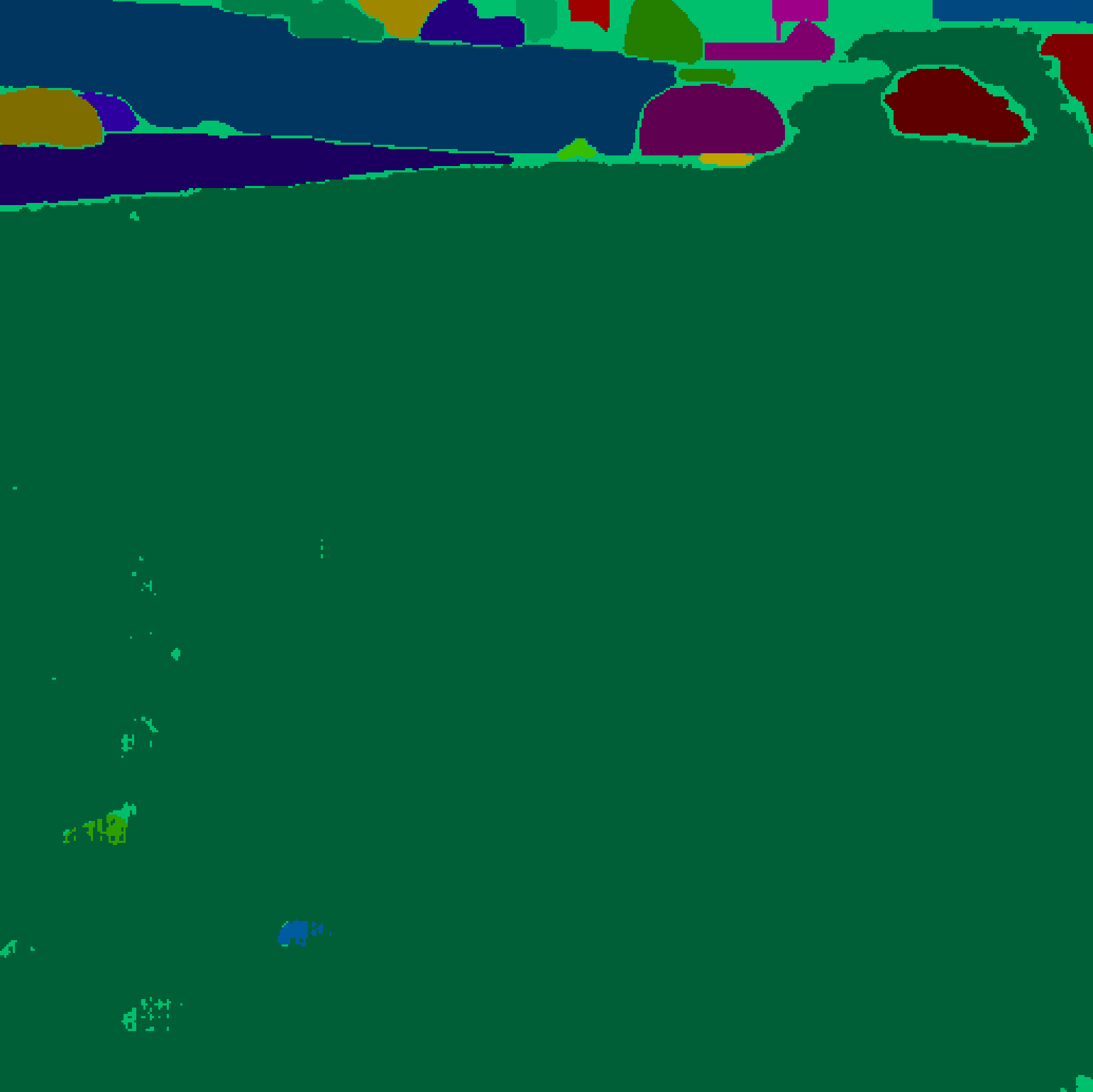} &
    \includegraphics[width=0.2\textwidth]{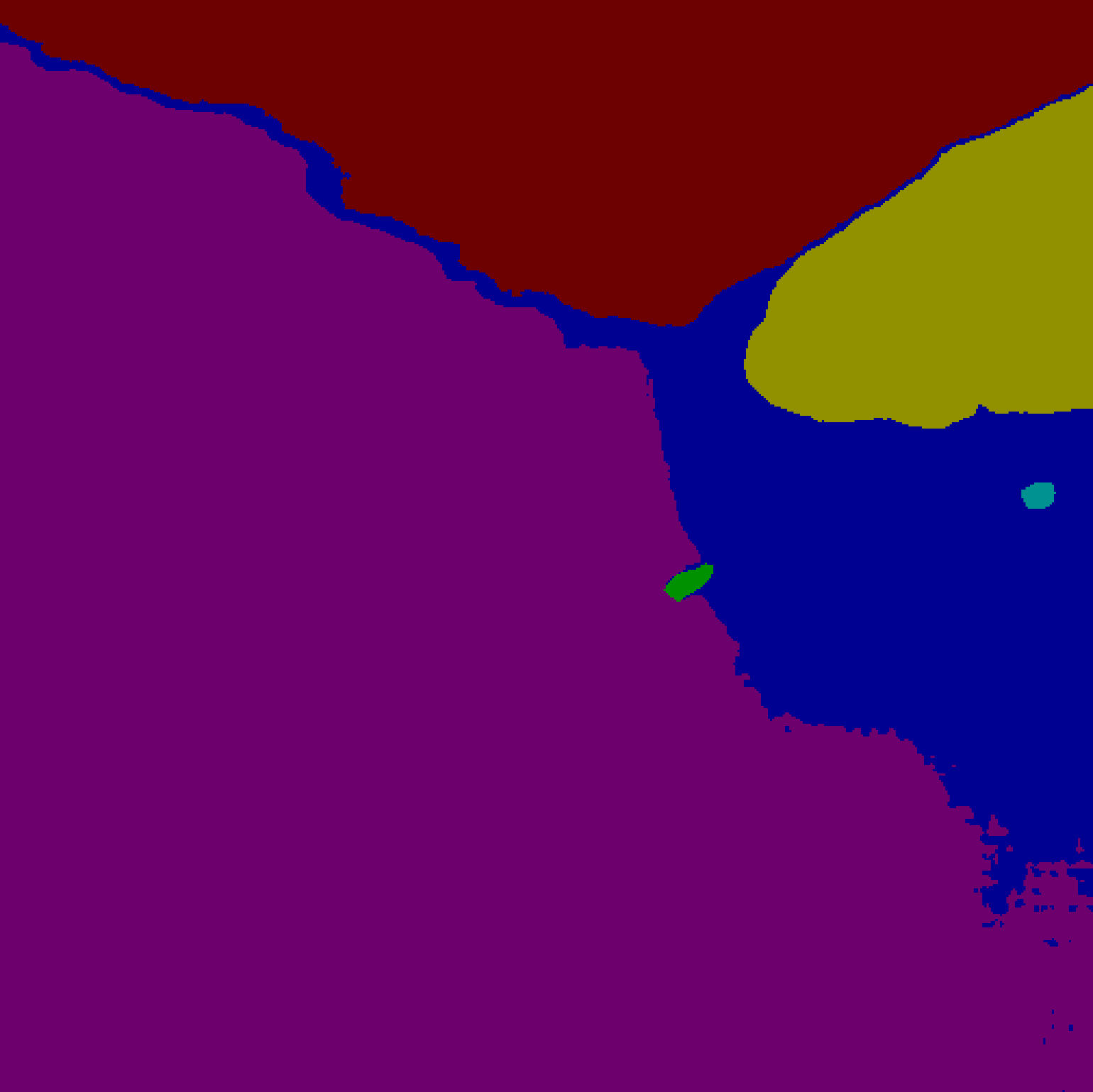} &
    \includegraphics[width=0.2\textwidth]{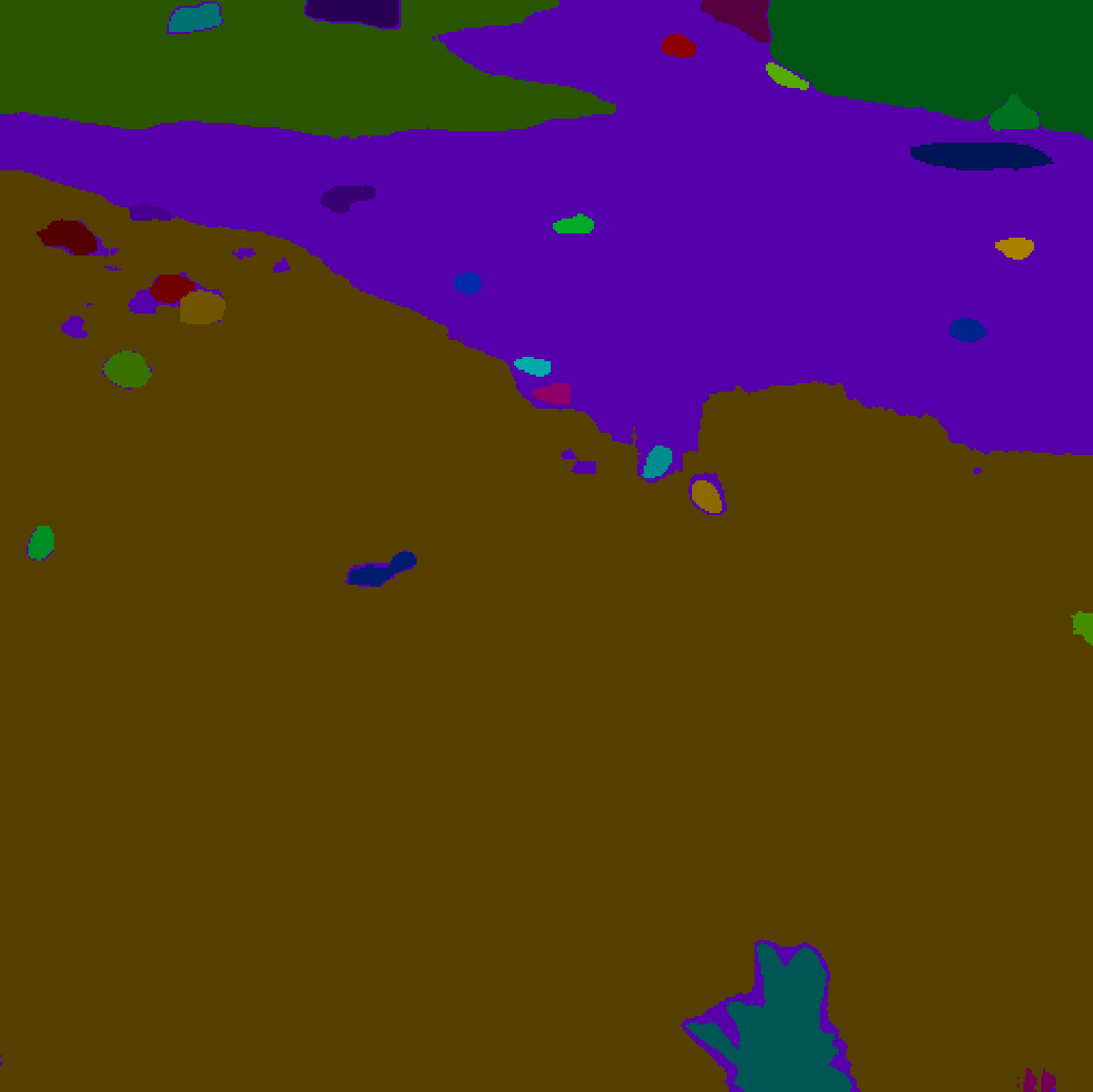} &
    \includegraphics[width=0.2\textwidth]{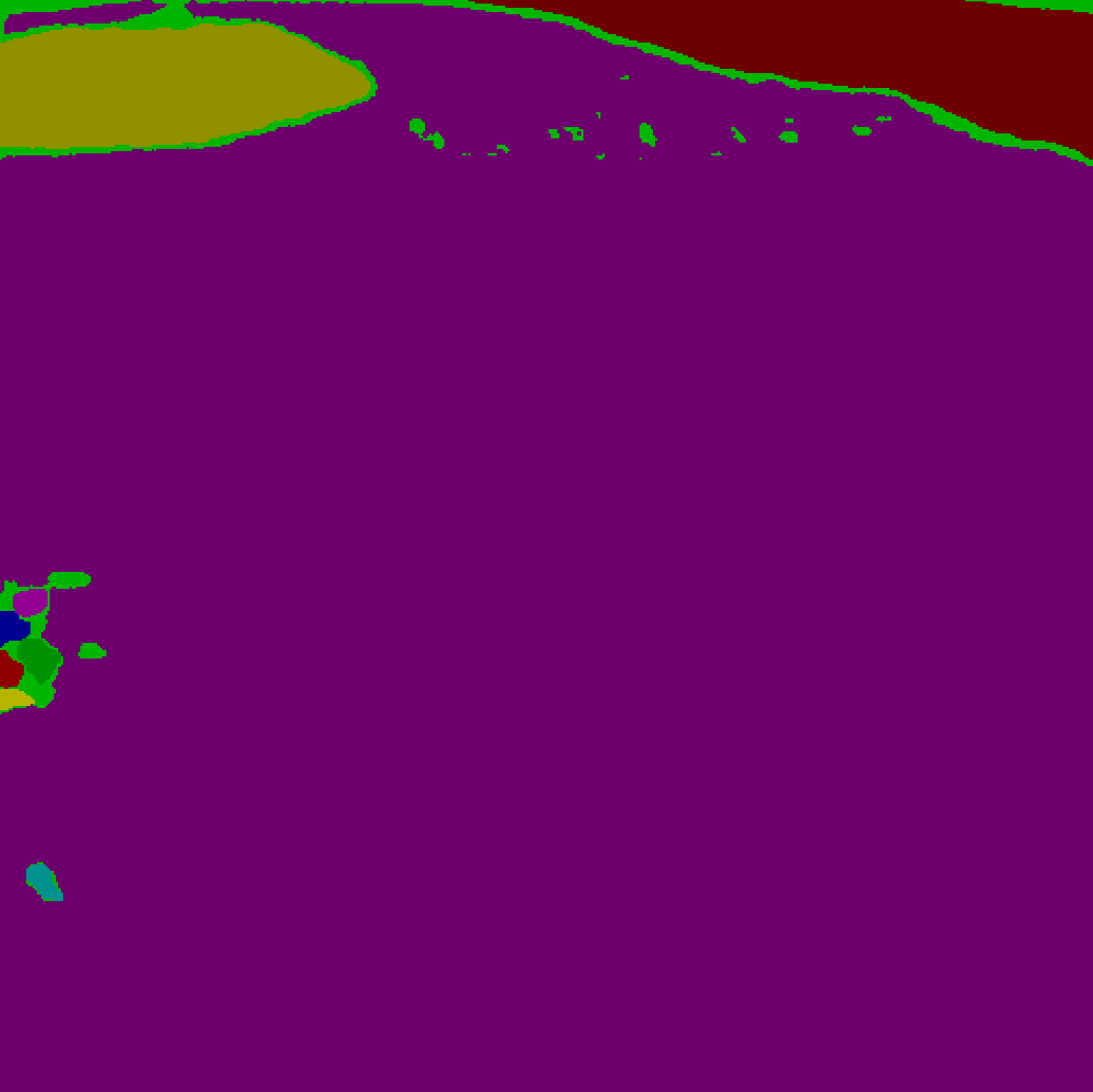} &
    \includegraphics[width=0.2\textwidth]{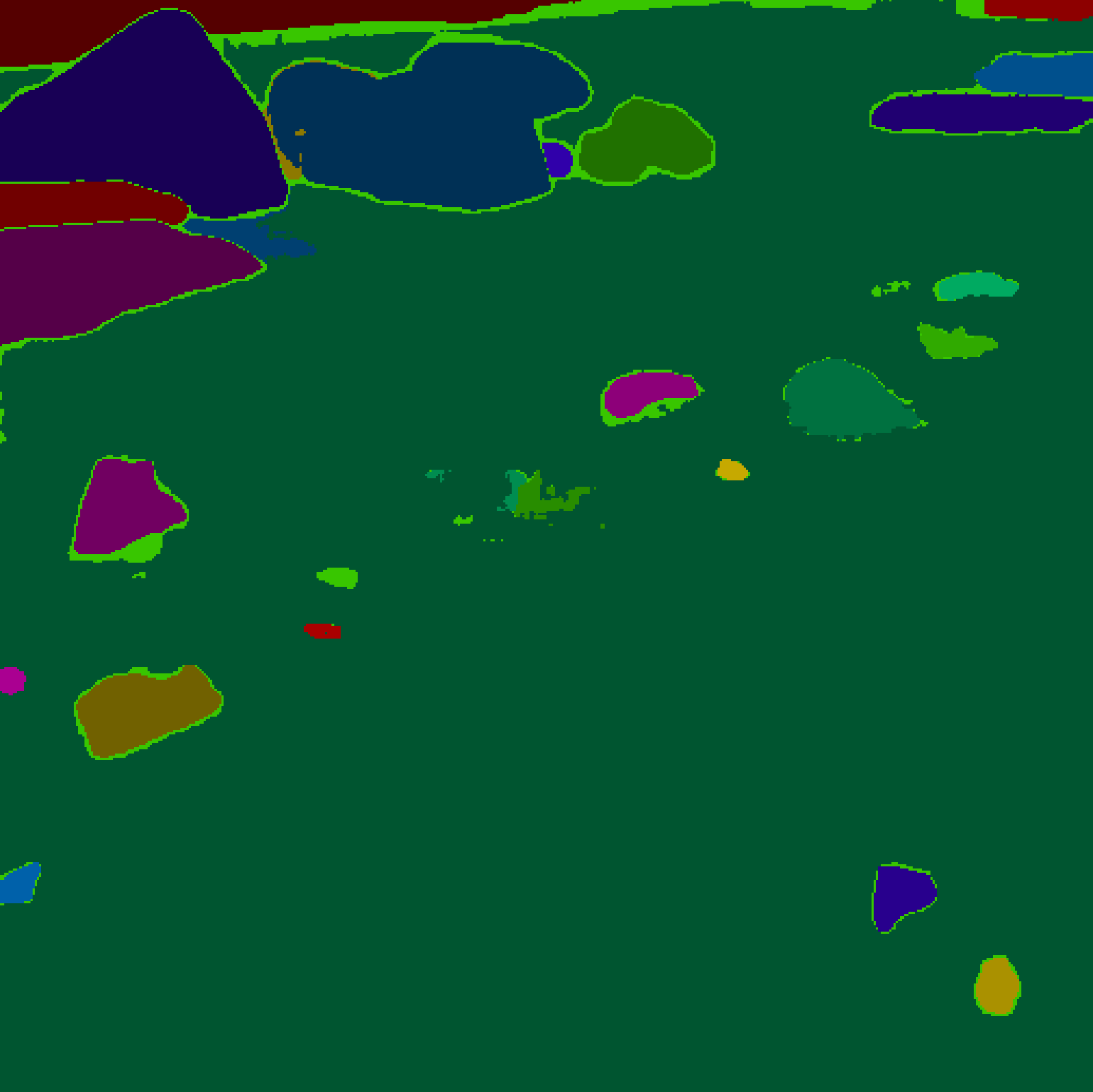} &
    \includegraphics[width=0.2\textwidth]{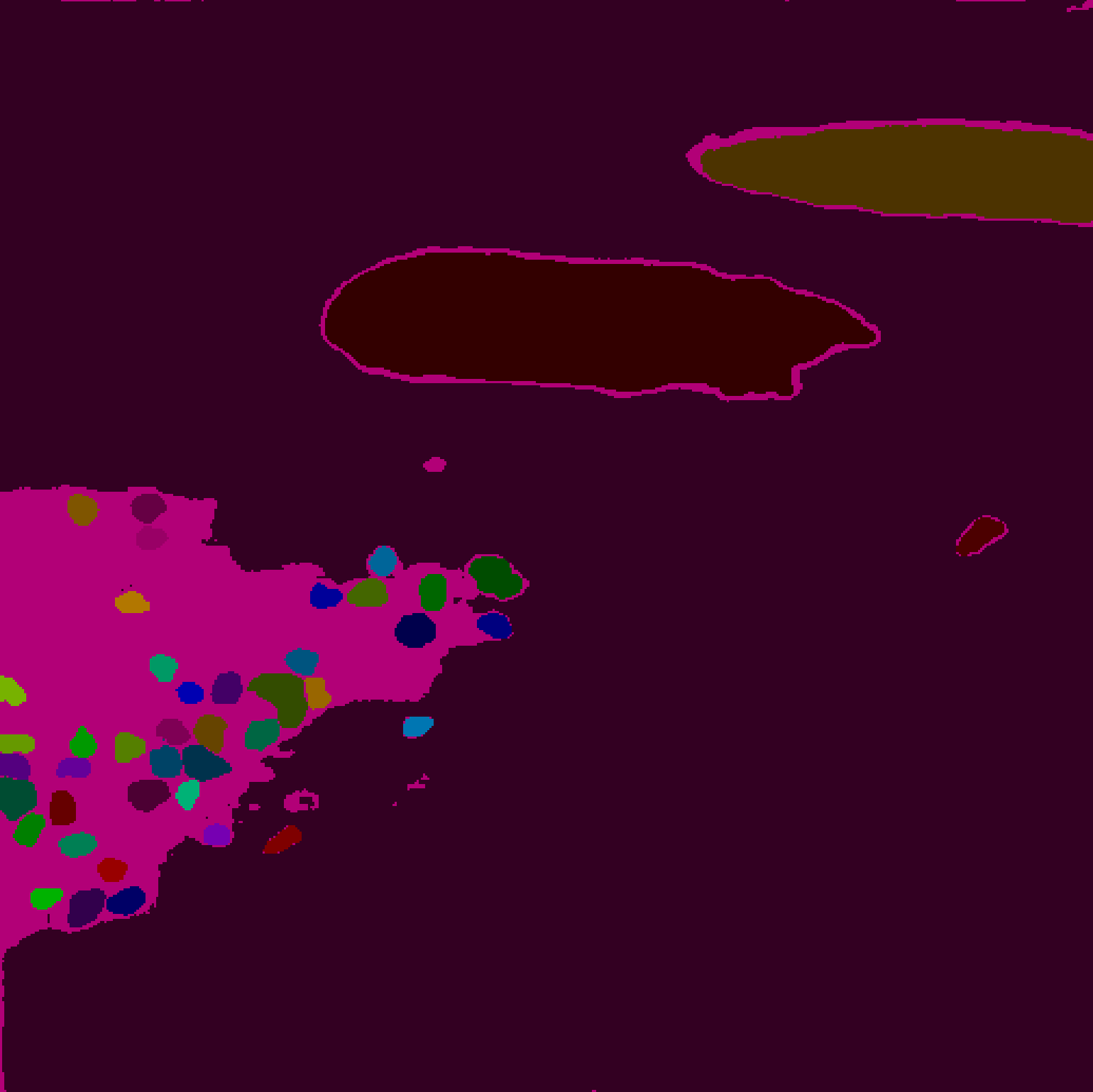} \\
    \rotatebox{90}{\parbox{4.0cm}{\centering \textbf{\model (RUGD)}}} &
    \includegraphics[width=0.2\textwidth]{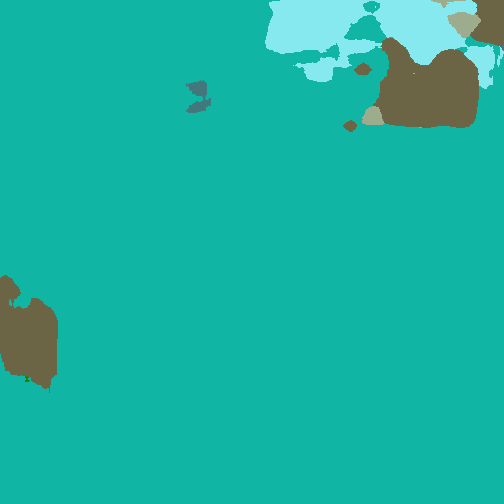} &
    \includegraphics[width=0.2\textwidth]{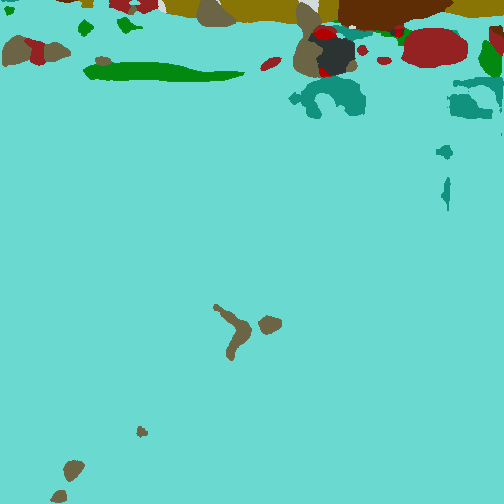} &
    \includegraphics[width=0.2\textwidth]{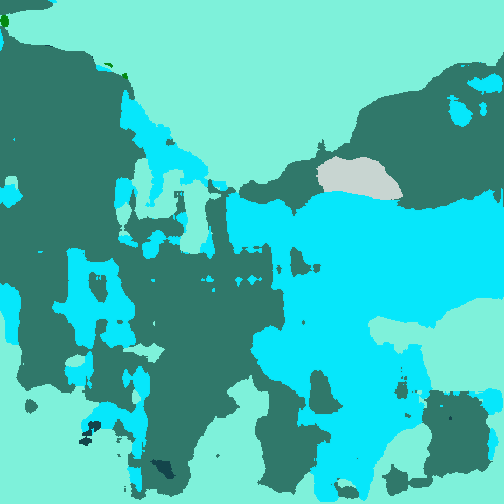} &
    \includegraphics[width=0.2\textwidth]{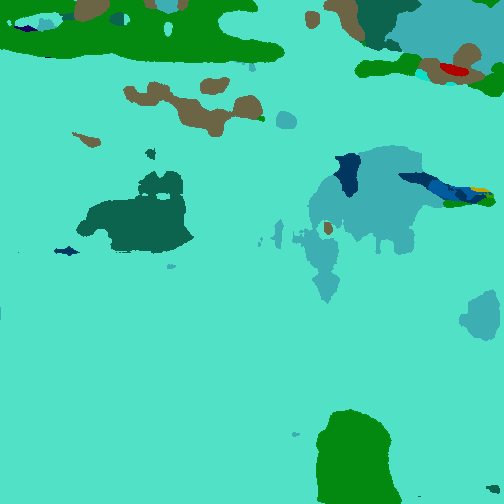} &
    \includegraphics[width=0.2\textwidth]{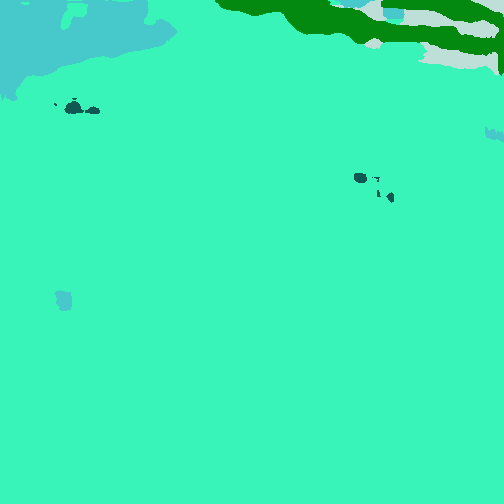} &
    \includegraphics[width=0.2\textwidth]{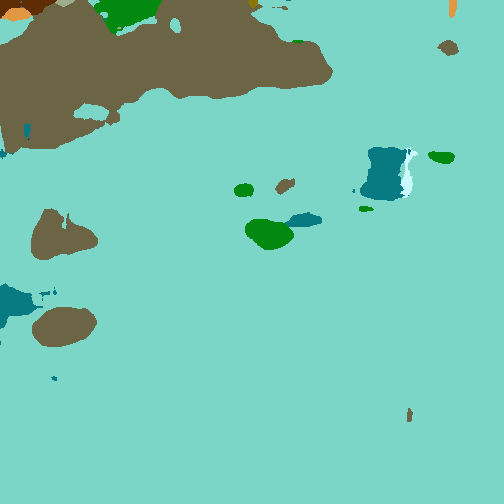} &
    \includegraphics[width=0.2\textwidth]{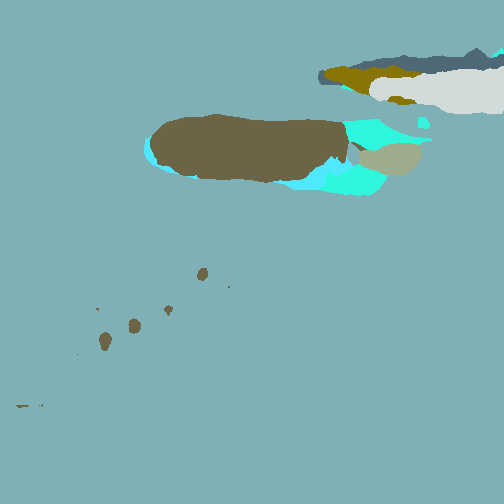} \\
\end{tabular}
}

\caption{Selected qualitative results for the EXTerra dataset. The coloring scheme follows that of \cref{fig:predictions_rugd}.}
\vspace{-1.5em}
\label{fig:predictions_dlr}
\end{figure*}

In this section we are looking into the relevance of synthetic data for the generalization capabilities of our model.
We train \model~with exclusively either the RUGD data or RUGDSynth, and show the results in~\cref{table:synth_real_ca_results_dlr} and~\cref{fig:predictions_dlr}.
The model trained on only synthetic data shows stronger metrics than the model only trained on RUGD, indicating the benfecial value of RUGDSynth.  
The union of both datasets shows the best results, which demonstrates that both can be used well together to fill each others shortcomings.

\begin{table}[!b]
\centering
\vspace{-1.2em}
\caption{Evaluation on EXTerra Dataset}
\vspace{-1.5em}
\resizebox{1\linewidth}{!}{
\begin{tabular}[t]{lS[table-format=2.2]S[table-format=2.2]S[table-format=2.2]S[table-format=2.2]}
\toprule
\multirow{2}{*}{Model} 
  & \multicolumn{1}{c}{cs} 
  & \multicolumn{3}{c}{ca} \\
\cmidrule(lr){2-2}
\cmidrule(lr){3-5}
 & mIoU & mIoU & \multicolumn{1}{c}{mPre} & \multicolumn{1}{c}{mRec} \\ 
\midrule
\vspace{0.5em}SAM\cite{kirillov_segment_2023} & \textbf{35.06} & 10.06 & \textbf{51.73} & 9.33  \\
Trinity (ours) &  &  &  &  \\
\Branch{RUGDSynth} & 29.78 & 34.93 & 48.67 & 28.04 \\
\Branch{RUGD} & 23.36 & 27. & 41.44 & 22.54 \\
\LastBranch{RUGDSynth+RUGD} & 34.82 & \textbf{41.84} & 51.57 & \textbf{34.82}\\
\bottomrule
\end{tabular}
}
\label{table:synth_real_ca_results_dlr}
\end{table}

\section{CONCLUSIONS}
We presented \model, a unified transformer-based architecture enabling flexible scene understanding without relying on fixed terrain taxonomy.
By introducing a novel split-transformer design with interacting query sets, our approach decouples robot-agnostic terrain representation from mission-relevant semantic prediction.
To support training at scale, we extended the OAISYS simulator and introduced RUGDSynth, a large synthetic dataset targeted for the proposed task.
Furthermore, we introduced EXTerra, a real-world planetary exploration dataset, showcasing the domain-shift capabilities of \model.
We hope that the proposed task formulation, datasets, and methodology will stimulate further research in this field and beyond and contribute to advancing the level of autonomy in future robotic systems.

\vspace{-0.2em}
\bibliographystyle{IEEEtran}
\bibliography{refs_short}

\end{document}